\documentclass[10pt,twocolumn,letterpaper]{article}

\usepackage{iccv}
\usepackage{times}

\usepackage{epsfig}
\usepackage{graphicx}
\usepackage[small]{caption}

\usepackage{amsmath}
\usepackage{nicefrac}

\usepackage{arydshln}
\usepackage{enumitem}
\usepackage[normalem]{ulem}

\usepackage{array}
\usepackage{booktabs} %
\usepackage{multirow} 
\usepackage{threeparttable}

\usepackage{color}
\usepackage{pifont}
\usepackage{ulem}

\usepackage{xspace}
\usepackage[table]{xcolor}

\usepackage{rotating}

\usepackage{enumitem}    
\usepackage{enumerate}

\usepackage[linesnumbered,ruled]{algorithm2e}

\usepackage{bm}

\usepackage{siunitx} %

\usepackage[american]{babel}
\usepackage{xcolor}
\colorlet{dark-blue}{blue!50!black}
\colorlet{dark-cyan}{cyan!75!black}
\colorlet{dark-purple}{purple!50!black}
\colorlet{dark-red}{red!75!black}
\colorlet{dark-green}{green!80!black}
\colorlet{dark-orange}{orange!50!black}
\colorlet{dark-gray}{black!75}
\colorlet{light-gray}{black!30}
\definecolor{nice-red}{HTML}{E41A1C}
\definecolor{nice-orange}{HTML}{FF7F00}
\definecolor{nice-yellow}{HTML}{FFC020}
\definecolor{nice-green}{HTML}{39b54a}
\definecolor{nice-blue}{HTML}{0071bc}
\definecolor{nice-purple}{HTML}{984EA3}

\definecolor{disco-d}{HTML}{E74C3C}
\definecolor{disco-i}{HTML}{9B59B6}
\definecolor{disco-s}{HTML}{3498DB}
\definecolor{disco-c}{HTML}{2ECC71}
\definecolor{disco-o}{HTML}{F1C40F}

\definecolor{darkgreen}{HTML}{3F7D31}
\definecolor{darkred}{HTML}{BA3132}

\definecolor{second}{rgb}{1, 0.85, 0.7}
\definecolor{best}{rgb}{1, 0.7, 0.7}
\definecolor{third}{rgb}{1,1, 0.8}

\colorlet{verylight-gray}{black!10}
\definecolor{LightCyan}{rgb}{0.66,0.85,0.76}

\makeatletter
\DeclareRobustCommand\onedot{\futurelet\@let@token\@onedot}
\def\@onedot{\ifx\@let@token.\else.\null\fi\xspace}

\def\eg{\emph{e.g}\onedot} 
\def\ie{\emph{i.e}\onedot}

\def\etal{\emph{et al}\onedot}
\makeatother

\newcommand{\figref}[1]{Figure~\ref{fig:#1}}%
\newcommand{\tabref}[1]{Table~\ref{tab:#1}} %
\newcommand{\eqnref}[1]{Equation~(\ref{eq:#1})}

\newcommand{\tb}[1]{\textbf{#1}}

\newcommand{\ignore}[1]{}   %

\newcommand{\revise}[1]{{#1}}

\newcommand{\cmark}{\textcolor{darkgreen}{\ding{51}}}
\newcommand{\xmark}{\textcolor{darkred}{\ding{55}}}

\newcommand{\setarial}[1]{
#1
}

\usepackage{xcolor}
\usepackage{amsmath}
\usepackage{mathtools}%

\usepackage[export]{adjustbox}

\makeatletter
\newcommand{\providelength}[1]{%
  \@ifundefined{\expandafter\@gobble\string#1}
   {%
    \typeout{\string\providelength: making new length \string#1}%
    \newlength{#1}%
   }
   {%
    \sdaau@checkforlength{#1}%
   }%
}

\newcommand{\sdaau@checkforlength}[1]{%
  \edef\sdaau@temp{\expandafter\sdaau@getfive\meaning#1TTTTT$}%
  \ifx\sdaau@temp\sdaau@skipstring
    \typeout{\string\providelength: \string#1 already a length}%
  \else
    \@latex@error
      {\string#1 illegal in \string\providelength}
      {\string#1 is defined, but not with \string\newlength}%
  \fi
}
\def\sdaau@getfive#1#2#3#4#5#6${#1#2#3#4#5}
\edef\sdaau@skipstring{\string\skip}
\makeatother

\DeclareMathOperator*{\argmin}{arg\,min}

\newcommand{\ba}{\mathbf{q}}
\newcommand{\bb}{\mathbf{r}}

\usepackage{makecell}

\usepackage{dsfont}
\usepackage{comment}

\graphicspath{{figure/}}

\usepackage[pagebackref=true,breaklinks=true,letterpaper=true,colorlinks,urlcolor=nice-red,citecolor=nice-green,linkcolor=nice-red,bookmarks=false]{hyperref}

\makeatletter
\def\@fnsymbol#1{\ensuremath{\ifcase#1\or \dagger\or \ddagger\or
		\mathsection\or \mathparagraph\or \|\or **\or \dagger\dagger
		\or \ddagger\ddagger \else\@ctrerr\fi}}
\makeatother

\iccvfinalcopy %
\def\iccvPaperID{} %

\usepackage[capitalize]{cleveref}
\crefname{section}{Sec.}{Secs.}
\Crefname{section}{Section}{Sections}
\Crefname{table}{Table}{Tables}
\crefname{table}{Tab.}{Tabs.}

\begin{document}

\title{\textcolor{disco-d}{D}\textcolor{disco-i}{i}\textcolor{disco-s}{s}\textcolor{disco-c}{C}\textcolor{disco-o}{O}: Portrait Distortion Correction with Perspective-Aware 3D GANs}

\author{Zhixiang Wang$^{1,2}$\footnotemark[1]\hspace{0.15in} 
    Yu-Lun Liu$^{3}$\hspace{0.15in}
    Jia-Bin Huang$^{4}$\hspace{0.15in}  
    Shin'ichi Satoh$^{2,1}$\\
    Sizhuo Ma$^{5}$\hspace{0.15in} 
    Gurunandan Krishnan$^{5}$\hspace{0.15in}  
    Jian Wang$^{5}$\footnotemark[2] \vspace{2mm}\\
    \hspace{0.1in} $^{1}$The University of Tokyo
    \hspace{0.10in} $^{2}$National Institute of Informatics\\
     $^{3}$National Yang Ming Chiao Tung University
    \hspace{0.10in} $^{4}$University of Maryland, College Park
    \hspace{0.10in} $^{5}$Snap Inc.\\
    \vspace{-2mm}\\
    \url{https://portrait-disco.github.io/}
    \vspace{-2mm}\\
}

\twocolumn[{%
\renewcommand\twocolumn[1][]{#1}%
\maketitle

\def\videopath{supplementary materials}
\ificcvfinal
\def\videopath{website~}
\fi

\vspace{-2em}
\begin{center}
    \begin{minipage}[c]{0.02\textwidth}
    \rotatebox[origin = c]{90}{Input images}
    \end{minipage}
    \begin{minipage}[c]{0.975\textwidth}
    \centering
        \fbox{\includegraphics[height=0.2\textwidth]{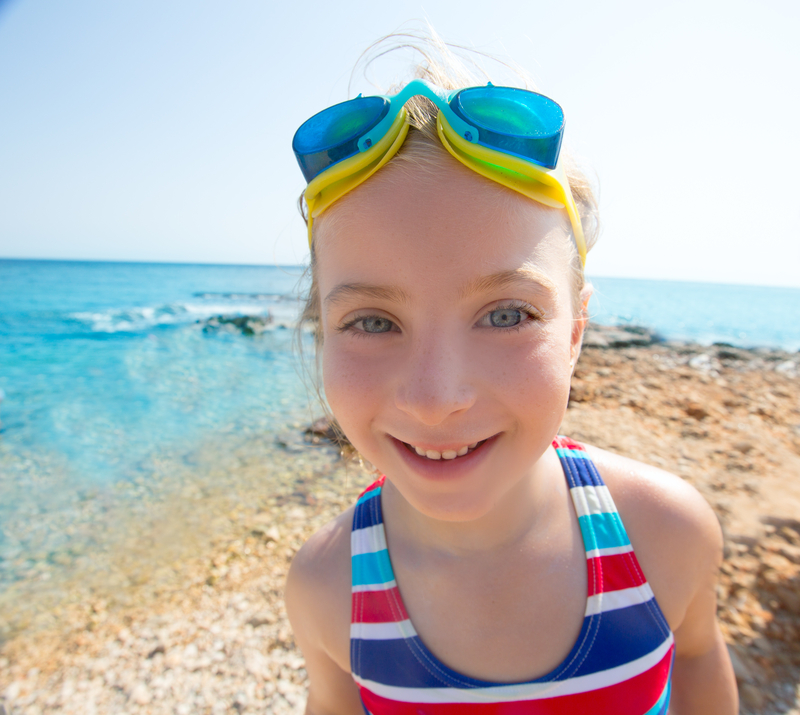}} \hfill
        \fbox{\includegraphics[height=0.2\textwidth]{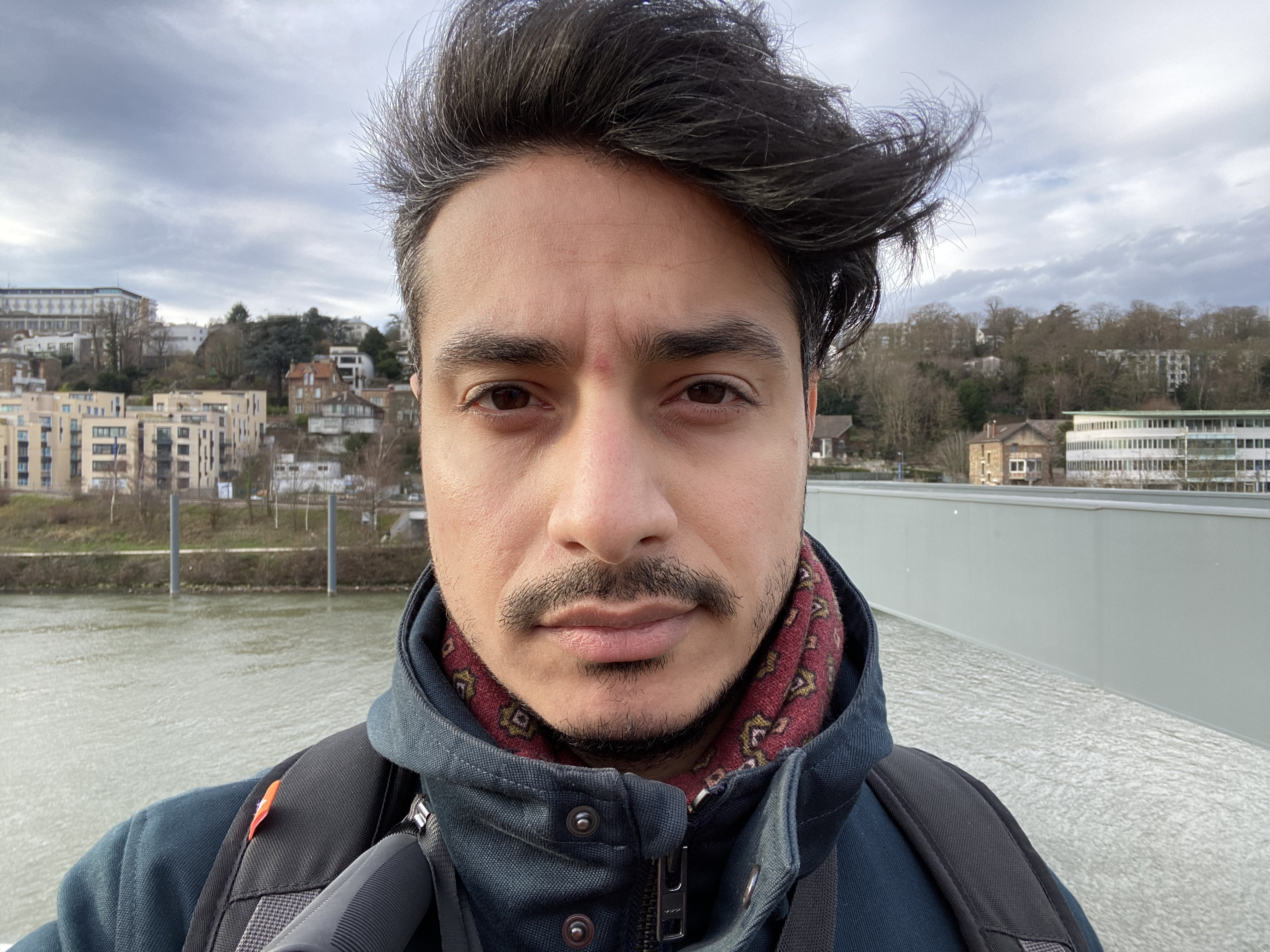}} \hfill        
        \fbox{\adjincludegraphics[trim={0 {.1\width} 0 {.4\width}},clip,height=0.2\textwidth]{dolly-zoom-out/np9-EYBFBI_input.jpg}} \hfill
        \fbox{\adjincludegraphics[trim={{.1\width} 0 {.1\width} 0},clip,height=0.2\textwidth]{dolly-zoom-out/np16-SPKCNT_input.jpg}}
    \end{minipage}
    \\
    \vspace{1mm}
    \begin{minipage}[c]{0.02\textwidth}
    \rotatebox[origin = c]{90}{Our results}
    \end{minipage}
    \begin{minipage}[c]{0.975\textwidth}
    \centering
        \fbox{\includegraphics[height=0.2\textwidth]{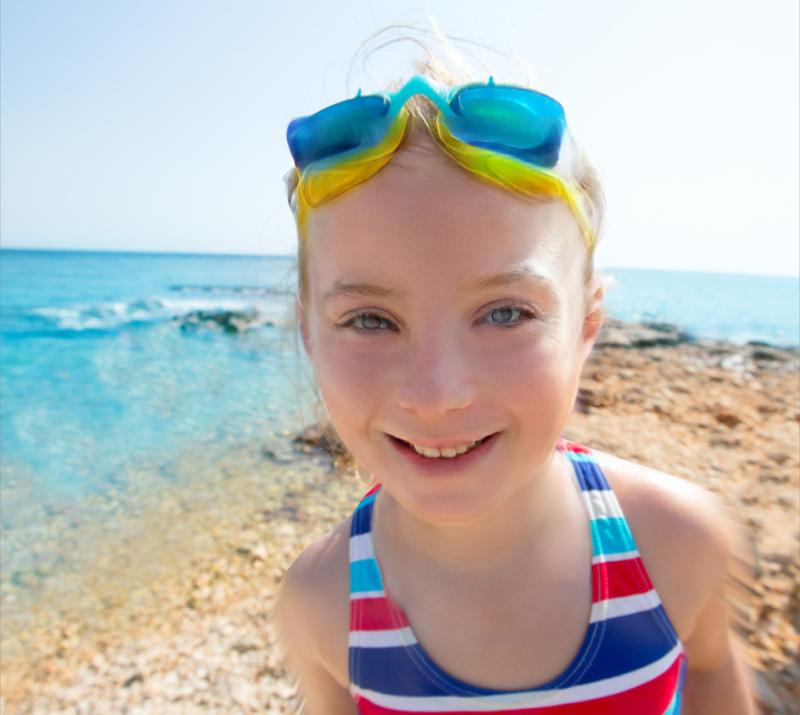}} \hfill  
        \fbox{\includegraphics[height=0.2\textwidth]{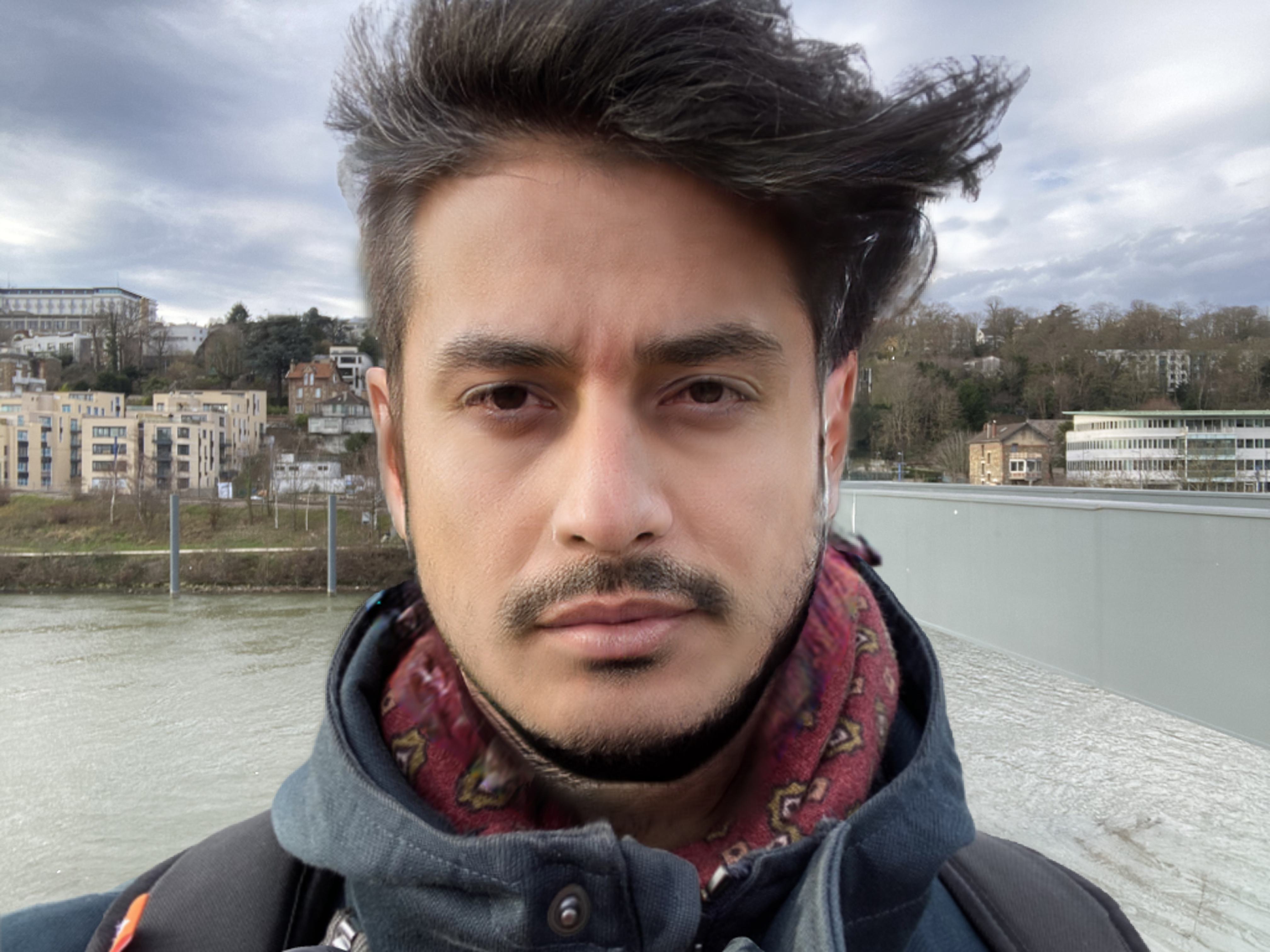}} \hfill
        \fbox{\adjincludegraphics[trim={0 {.1\width} 0 {.4\width}},clip,height=0.2\textwidth]{dolly-zoom-out/np9-EYBFBI_oursx4.jpg}} \hfill
        \fbox{\adjincludegraphics[trim={{.1\width} 0 {.1\width} 0},clip,height=0.2\textwidth]{dolly-zoom-out/np16-SPKCNT_oursx2.jpeg}}
    \end{minipage}
    \vspace{-2mm}
\captionof{figure}{
\tb{Portrait distortion correction.} 
Portrait photos captured from a short distance (\eg, selfie) often suffer from undesired perspective distortions (the first row). 
Our approach corrects these perspective distortions and synthesizes visually pleasant views by \emph{virtually} enlarging the focal length and moving the camera further away from the subject. Please check the \videopath for videos.
}
    \label{fig:teaser}
\end{center}
}]

\ificcvfinal
\renewcommand{\thefootnote}{\fnsymbol{footnote}}
\footnotetext[1]{Part of the work was done 
while Zhixiang was 
an intern at Snap Research, NYC.}
\footnotetext[2]{Corresponding author}
\fi

\maketitle

\begin{abstract}
\vspace{-4mm}
Close-up facial images captured at short distances often suffer from perspective distortion, resulting in exaggerated facial features and unnatural/unattractive appearances. 
We propose a simple yet effective method for correcting perspective distortions in a single close-up face.
We first perform GAN inversion using a perspective-distorted input facial image by jointly optimizing the camera intrinsic/extrinsic parameters and face latent code. 
To address the ambiguity of joint optimization, we develop starting from a short distance,
optimization scheduling, reparametrizations, and geometric regularization. 
Re-rendering the portrait at a proper focal length and camera distance effectively corrects perspective distortions and produces more natural-looking results. 
Our experiments show that our method compares favorably against previous approaches qualitatively and quantitatively. %
We showcase numerous examples validating the applicability of our method on \textbf{in-the-wild} portrait photos. 
We will release our code and the evaluation protocol to facilitate future work.
\end{abstract}

\section{Introduction}
\label{sec:introduction}

Every day, millions of people enjoy taking selfies with their smartphones. 
Although these devices have high-quality cameras that can capture high-resolution and accurate colors, selfies tend to suffer from perspective distortion.
This distortion is caused by the short distance between the face and the camera (usually between 20--60 cm) and is particularly noticeable (as shown in the first-row of~\figref{teaser}). 
The distortion makes frontal features, like the nose, appear more prominent and causes the face to look unnatural and asymmetrical. 
Additionally, the distortion often obscures the side of the face, including the ears.
This distortion creates unflattering images and could negatively impact face identification and other related tasks.

\begin{figure*}[t]

\small
\centering
\begin{tabular}{
    @{\hspace{0mm}}c@{\hspace{1.8mm}} 
    @{\hspace{0mm}}c@{\hspace{1.8mm}} 
    @{\hspace{0mm}}c@{\hspace{1.8mm}}
    @{\hspace{0mm}}c@{\hspace{1.8mm}}
}
\fbox{\includegraphics[height=0.18\linewidth]{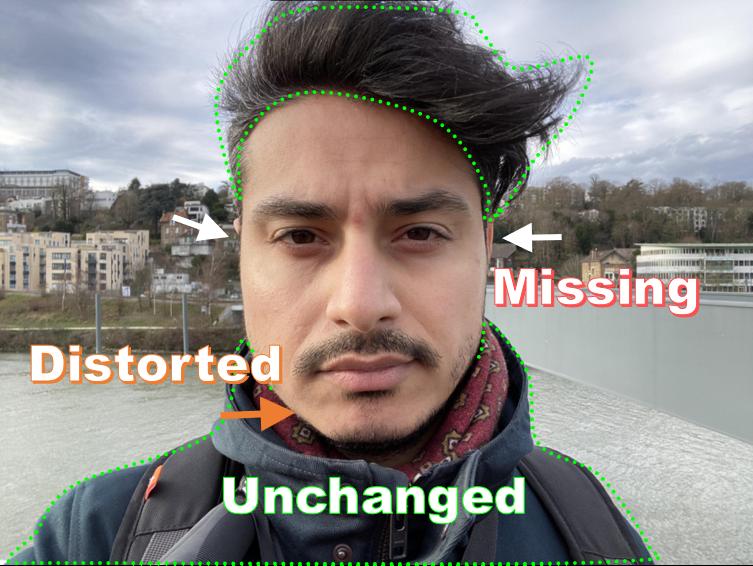}} & 
\fbox{\includegraphics[height=0.18\linewidth]{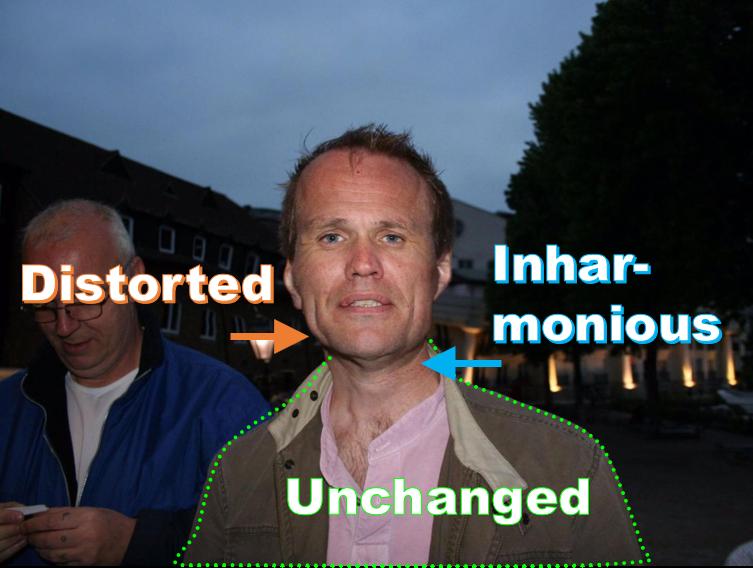}} &
\fbox{\includegraphics[height=0.18\linewidth]{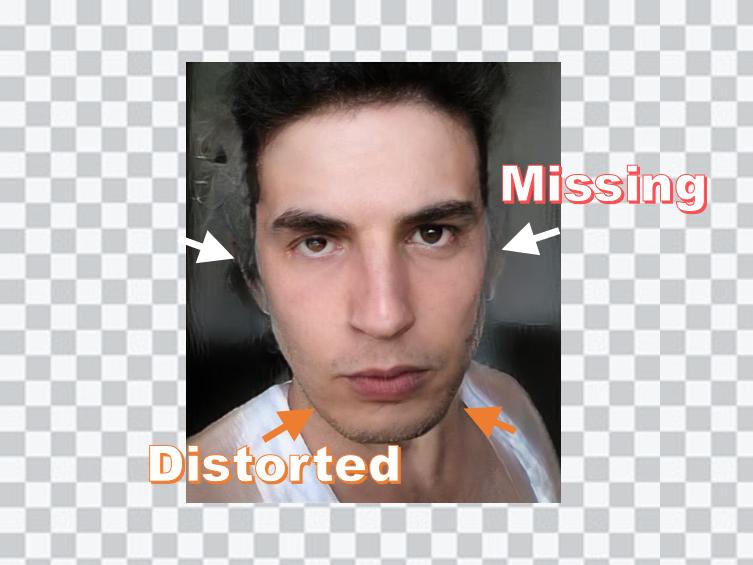}} &
\fbox{\includegraphics[height=0.18\linewidth]{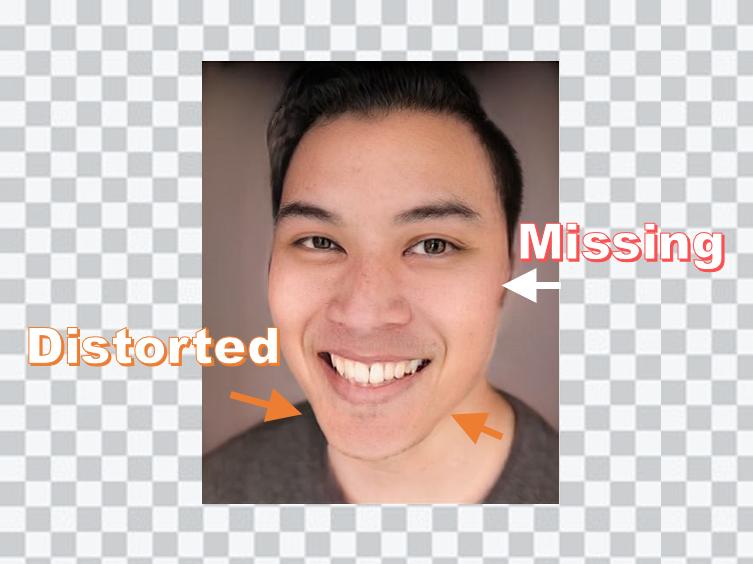}} \\

(a) Fried's~\cite{fried2016perspective} & (b) Zhao's \cite{Zhao-2019-ICCV} & (c) PTI~\cite{Roich-2021-TOG-PTI} & (d) Ko's~\cite{ko20233d} \\
\end{tabular}
\vspace{-2mm}
\caption{
\textbf{Limitations of state-of-the-art portrait perspective correction techniques.}
{
(a)(b)~\cite{fried2016perspective} and \cite{Zhao-2019-ICCV} are 2D warping-based methods that cannot fully recover the correct face geometry or generate missing content, such as ears. Moreover, (b) shows that the corrected image using \cite{Zhao-2019-ICCV} exhibits an inharmonious composition of the face and neck, in contrast to our result in \figref{full-frame}.
(c)(d) are GAN inversion methods that can manipulate camera parameters.
(c) PTI~\cite{Roich-2021-TOG-PTI} is a 2D GAN inversion method that may produce sub-optimal solutions and incorrect facial geometry when applied to 3D GANs. (d) is a 3D GAN inversion method that jointly optimizes face and partial camera parameters but cannot generate correct geometry. Both (c) and (d) can only correct facial regions instead of the full body.
} 
}
\vskip -8pt
\label{fig:traditional_cropping}
\end{figure*}

Existing efforts automatically correct portrait perspective distortions~\cite{beeler2010high,bryan2012perspective,burgos2014distance} often involving reconstruction-based warping~\cite{fried2016perspective} and learning-based warping~\cite{Zhao-2019-ICCV,nagano2019deep}. 
However, these methods rely on estimating a 2D flow map to warp the image, leading to incorrect face shapes after correction, as shown in \figref{traditional_cropping}(a). Moreover, they cannot generate disoccluded pixels, such as ears and hairs, which may be revealed in the background. Additionally, the warping-based method cannot render the background with the same camera parameters, causing misalignment between the face and body.

Our proposed solution to correct portrait perspective distortion is \emph{3D GAN inversion}, building on the effectiveness of 3D GANs~\cite{niemeyer2021giraffe,zhou2021cips,chan2021pi,or2022stylesdf,chan2022efficient,sun2022ide,deng2022gram}. 
This approach optimizes facial latent code, camera pose, and focal length to estimate facial geometry and camera-to-face distances. 
However, optimizing these parameters from a single distorted face is challenging, and existing GAN inversion methods like PTI~\cite{Roich-2021-TOG-PTI} fail to provide accurate results when applied to 3D GANs. 
To address this issue, we propose four designs:
(1) closeup camera-to-face distance initialization,
(2) separate optimization of face and camera parameters, 
(3) reparameterizations, and  
(4) landmark and geometric constraints.
We also incorporate a workflow to handle 
 full images rather than cropped faces. 
Our method can correct perspective distortion by adjusting the camera-to-face distance (as shown in the second row of~\figref{teaser}) and applying special visual effects such as dolly-zoom by adjusting camera parameters.

We make the following contributions: 
\begin{itemize}
\item We propose a pipeline for correcting portrait distortion using perspective-aware 3D GAN inversion. 
 Our pipeline integrates GAN inversion for the face region and a workflow to achieve camera-consistent full-image manipulation, avoiding inharmonious composition between the face and body. This enables various visual effects, including dolly-zoom videos. 
\item We explore several design choices to avoid the optimization falling into sub-optimal solutions, including better initialization, separate optimization of face and camera parameters, reparameterizations, and geometric loss.
\item We establish a comprehensive evaluation for portrait perspective distortion correction, including quantitative, qualitative, full-image, and video evaluation, which will benefit future research in this area. 
\end{itemize}

\section{Related Work}
\label{sec:related}

\subsection{Portrait perspective undistortion}

Selfie photos taken from close distances often suffer from perspective distortions, resulting in unappealing distortions such as an enlarged nose, uneven facial features, asymmetry, and hidden ears and hairs. 
These distortions are commonly referred to as ``selfie effects" and are a significant concern for many people, with some even considering plastic surgery as a solution \cite{ward2018nasal}. 
Research indicates that the camera distance plays a vital role in portrait perception, and studies have identified an ``optimal distance" for capturing undistorted facial images~\cite{bryan2012perspective,cooper2012perceptual}. 
Specifically, it has been found that 50mm lenses are ideal for producing natural-looking and flattering images. 
In response, smartphone manufacturers have attempted to encourage users to take selfies from a greater distance by reducing the field of view \cite{williams2017camera}.

Current perspective distortion methods either model distortion as a warping function parameter~~\cite{valente2015perspective} or manipulate camera-to-face distance in a reconstructed model \cite{fried2016perspective}. 
While deep learning-based methods~\cite{Zhao-2019-ICCV} can correct minor distortions, they struggle with severe distortions due to inaccurate 3D face-fitting steps and the inability to inpaint occluded regions like ears using 2D warping flow maps.
3D radiance field-based methods \cite{gao2020portrait,athar2022rignerf,gafni2021dynamic} provide full control of camera parameters but require many training images and do not leverage face priors. 
Our method uses 3D GAN inversion to correct close-range input images, fill in unobserved regions, and allow flexible camera-to-face distances, effectively correcting severe distortions.

\subsection{3D GANs} The neural 3D representation~\cite{park2019deepsdf,michalkiewicz2019implicit,atzmon2019sal,gropp2020implicit,sitzmann2019srns,peng2020convolutional,mescheder2019occupancy,chen2019learning,mildenhall2020nerf,wang2021neus,mildenhall2020nerf}
has shown impressive photorealism in novel view synthesis and is a foundational representation for 3D-aware generation.
Implicit 3D representations have been leveraged by recently proposed 3D GANs~\cite{deng2022gram,chan2021pi,or2022stylesdf,zhou2021cips,chan2022efficient,niemeyer2021giraffe} to generate high-resolution outputs with remarkable details and 3D consistency.
Our work uses the pre-trained architecture in  EG3D~\cite{chan2022efficient} due to its computational efficiency and its ability to produce photorealistic 3D consistent images, similar to those generated by StyleGANs~\cite{karras2019style,karras2020analyzing}. 
However, our method is agnostic to the choice of 3D GANs.

\subsection{GAN inversion}
GAN inversion is a technique that maps a real image back into the latent space of a pre-trained GAN, which can expand the model's editing capability to real photos. 
There are two main categories of GAN inversion: 2D and 3D. 
2D GAN inversion methods optimize the latent code for a single image~\cite{abdal2019image2stylegan,creswell2018inverting} or use a learned encoder to project images to the latent space~\cite{richardson2021encoding,tov2021designing,alaluf2021restyle}. 
Some hybrid strategies combine both methods to refine the latent code by optimization~\cite{guan2020collaborative,zhu2016generative}.
Recent 2D GAN inversion methods achieve high editing capabilities and have been extended for video editing~\cite{xu2022temporally,tzaban2022stitch,abdal2022video2stylegan}. 
However, editing 3D-related attributes such as camera parameters and head pose remains inconsistent and prone to severe flickering, as the pre-trained generator is unaware of the 3D structure.

On the contrary, 3D GAN inversion methods~\cite{ko20233d,lin20223d,sun2022ide,wang2022narrate,xu2023n,xie2023high} achieve 3D consistent reconstruction and manipulation by incorporating 2D GAN inversion methods, such as PTI~\cite{Roich-2021-TOG-PTI}, with estimated camera parameters obtained from 3DMM or other algorithms. While some recent methods like \cite{lin20223d} and \cite{wang2022narrate} estimate all camera parameters from 3DMM and keep them fixed, Ko~\etal~\cite{ko20233d} assume known camera intrinsics and camera-to-face distances to jointly optimize the face latent code and rest of the camera parameters. However, correcting perspective distortion requires estimating the face latent code, camera-to-face distance, and focal length, posing a challenge due to ambiguity among these parameters. To address this, we propose a perspective-aware 3D GAN inversion method to estimate the face latent code and camera parameters accurately.

\section{\revise{Background}}
{ 
We will briefly introduce the basics of StyleGAN and StyleGAN inversion, followed by 3D GANs.}

\paragraph{StyleGAN}
{
Given a random sample $\mathbf{z}\!\in\!\mathds{R}^{512}$ drawn from a normal distribution, StyleGAN~\cite{karras2019style} can yield a new sample from the data distribution.
It first maps $\mathbf{z}$ to an intermediate latent vector $\mathbf{w} \in \mathds{R}^{512}$ using a learned mapping $\mathbf{w} = H_\theta(\mathbf{z})$. 
The space of the latent vector $\mathbf{w}$ (style code) is commonly referred to as $\mathcal{W}$.
The vector $\mathbf{w}$ controls feature normalization in 18 layers of the generator network $G_\theta$ and produces the final image 
\begin{equation}
    I = G_\theta(\mathbf{w}) = G_\theta(H_\theta(\mathbf{z}))\,.
\end{equation}
}

\begin{figure*}[!t]
    \centering
    \includegraphics[width=0.85\textwidth]{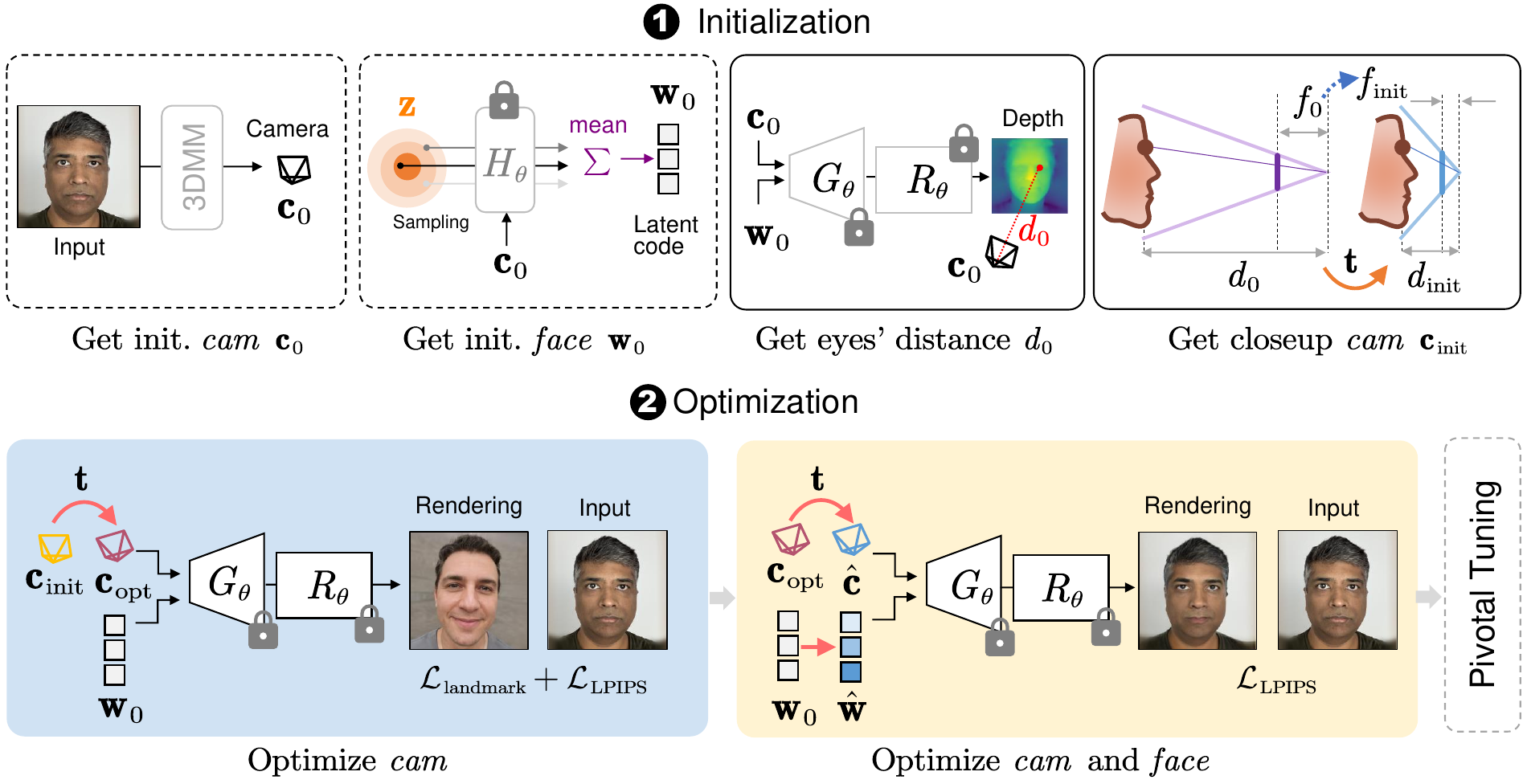}
    \vspace*{-1mm}
    \caption{
    \textbf{Perspective-aware 3D GAN inversion.}
    {
    \tb{Step 1}: \emph{Initialization}. We first fit a 3DMM model to the image to get an initial camera pose and average randomly sampled latent codes to initialize the face latent code. The initialized camera pose can roughly match the face direction and size, but the estimated focal length and camera-to-subjective distance are inaccurate.
    Then, we get a closeup camera by pushing the camera-to-face distance $d_0$ to a small value $d_\text{int}$ and changing the focal length according to the reparameterization method. 
    \tb{Step 2}: \emph{Optimization}. We fix the face latent code, generator, and neural renderer to optimize the camera parameters.
    Here, we reparameterize the focal length and rotation to further ease optimization.
    After optimizing the camera poses, we simultaneously optimize the face latent code and camera parameters. 
    Finally, we perform pivotal tuning to fine-tune the generator to achieve high-fidelity results on real images. 
    }
    }
    \label{fig:inver-method}
\end{figure*}

\paragraph{StyleGAN inversion} 
enables the projection of an input real image, denoted as $x$, into the pre-trained generator's domain. This projection allows us to perform various editing operations on the input image. Given the  exceptional fine-grained editing ability, inversion is typically carried out in the $\mathcal{W}$ space. 
To obtain the optimal latent vector $\hat{\mathbf{w}} \in \mathcal{W}$, we minimize the LPIPS perceptual loss function~\cite{zhang2018unreasonable}:
\begin{equation}\label{eq:objective}
    \hat{\mathbf{w}} = \argmin_{\mathbf{w}} \mathcal{L}_\text{LPIPS}(G_\theta(\mathbf{w}), x)\,.
\end{equation}
Due to potential disparities between the real image and the pre-trained generator's domain, the reconstructed image using the inverted latent code $\hat{\mathbf{w}}$ might suffer from distortion. To address this, Roich~\etal~\cite{Roich-2021-TOG-PTI} propose \emph{pivotal tuning} that unfreezes and fine-tunes the generator using fixed $\hat{\mathbf{w}}$.
The primary objective is to optimize the generator's parameters
\begin{equation}\label{eq:pt}
\begin{split}
    \vartheta = \argmin_{\theta} \mathcal{L}_\text{LPIPS}(G_\theta(\mathbf{w}), x)  + \lambda_{L2}\mathcal{L}_{L2}(G_\theta(\mathbf{w}), x)\,.
\end{split}
\end{equation}

\paragraph{3D GAN} 
{
combines the implicit 3D representation and StyleGAN for 3D controllable image generation.
The StyleGAN, including $H_\theta$ and $G_\theta$, uses latent codes and camera parameters as input to generate implicit 3D representation. Then, the neural renderer $R_\theta$ takes the implicit representation and camera parameters to produce the final image. 
The formulation of this process is given by:
\begin{equation}
    I = R_\theta(G_\theta(\mathbf{w}), \mathbf{c}) = R_\theta(G_\theta(H_\theta(\mathbf{z}, \mathbf{c})), \mathbf{c}),
    \label{eq:generator}
\end{equation}
where $\mathbf{c}$ includes the intrinsic and extrinsic parameters. 
}

\section{\revise{Perspective-aware 3D GAN Inversion}}
\label{sec:method}

{Correcting the perspective distortion of a single close-up face portrait requires manipulation of its camera-to-subject distance.
We propose a perspective-aware GAN inversion technique that utilizes pre-trained 3D GANs to invert the portrait into its corresponding face latent code and camera parameters  (see \figref{inver-method}). 
Then, we adjust the camera parameters, such as the camera-to-subject distance and focal length to re-render a novel portrait with alleviated distortion.

Existing methods~\cite{sun2022ide,lin20223d,ko20233d,xie2023high} extend  PTI to 3D GAN inversion by introducing additional camera parameters. However, the accuracy of these parameters, especially the focal length and camera-to-subjective distance, can be uncertain when estimated using 3DMM or other algorithms.
Nonetheless, these methods can still produce reasonable results despite the errors because their input images are captured at far distances, where the weak perspective model can be approximated, and the input reflects ground truth faces (see \figref{weak_perspective}). 
The inaccuracies in focal length and camera-to-subjective distance merely lead to minor scale discrepancies in the face geometry.

However, close-up photography is an \textbf{entirely different} story due to the perspective model, and the distortion that makes the face appearance differ from the ground-truth face (\figref{weak_perspective}).
Therefore, using these inaccurate parameters directly could lead to faces with incorrect geometry (see \figref{av}). 
For high-quality 3D face images, accurate estimation of \emph{both} camera-to-subject distance and focal length is essential. Therefore, we jointly optimize the camera parameters and the face latent code:
\begin{equation}\label{eq:objective2}
    \hat{\mathbf{w}}, \hat{\mathbf{c}} = \argmin_{\mathbf{w},\, \mathbf{c}} \mathcal{L}(R_\theta(G_\theta(\mathbf{w}), \mathbf{c}), x) \,.
\end{equation}

Inferring unknown face and camera parameters from a single image is indeed an \emph{ill-posed} problem, as there can be multiple combinations of focal length, camera-to-subject distance, and face shape that produce the input image (see \figref{ambiguity}). 
Due to this ambiguity, combing na\"ive camera optimization with PTI encounters significant challenges
(as shown in \figref{av}, \ref{fig:naiive}). To alleviate the ambiguity, we propose a perspective-aware 3D GAN inversion with four techniques: starting from a short distance, optimization scheduling, reparameterizations, and landmark regularization.

\begin{figure*}[!t]
    \centering
    \includegraphics[width=0.98\textwidth]{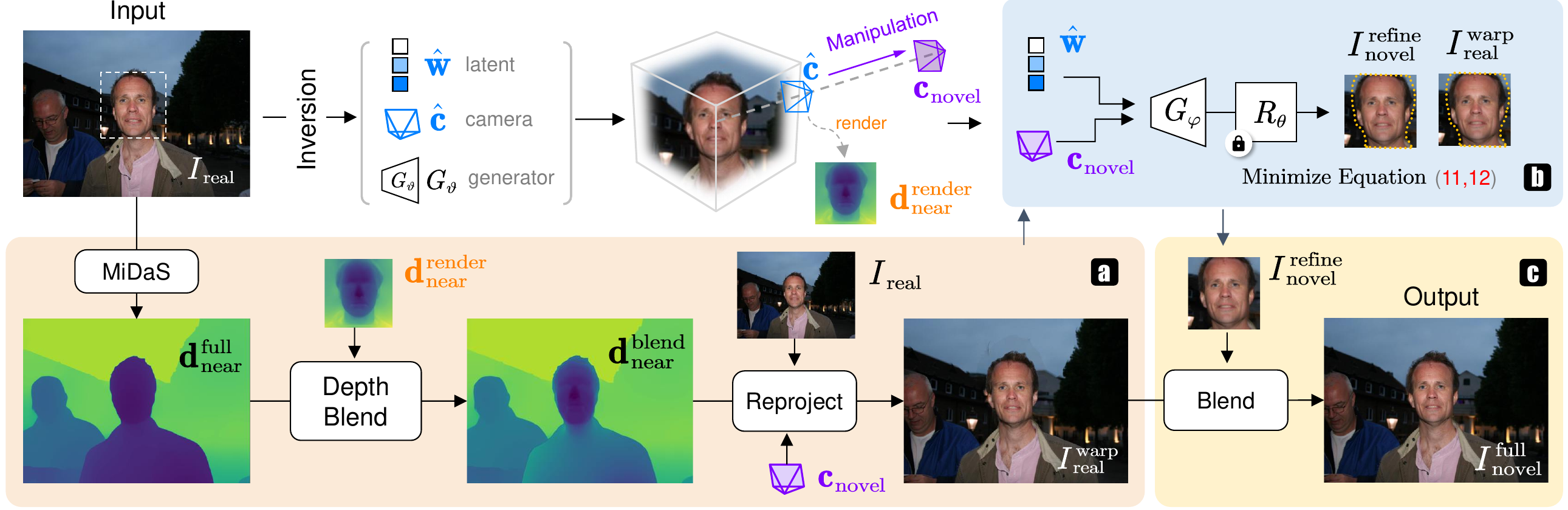}
    \vspace{-2mm}
    \caption{\textbf{
    \revise{Pipeline of processing full-frame image.}
    } 
    Taking a full-frame close-up face image, we crop the closest face from the input image and perform 3D GAN inversion to infer the face latent code and camera parameters of the cropped face. 
    After inversion, we manipulate the camera distance and focal length to render virtual images. (a-c) Geometry-aware stitching tuning.
    (a) We align and blend the rendered face depth map with the depth estimated from the entire image using a monocular depth estimation algorithm (MiDaS~\cite{ranftl2020towards}). 
    We project the entire input image to the same virtual camera positions of the manipulated face image. 
    (b) 
    We fine-tune the generator by minimizing border loss and content loss to refine the border of the generated long-distance image.  
    (c)
    Finally, we blend the warped full image with the generated face image.    
    }
    \label{fig:method}
\end{figure*}

\subsection{Initialization}

We tried to use a method similar to existing 3D GAN inversions for camera and face initialization. However, the initialized camera parameters are unsuitable for the desired setting, 
{where a close-up camera is required.}

\paragraph{Starting from a short distance}
Since the initialized camera $\mathbf{c}_0$ can generate a face match with the size of the face in the input image, we refine it to a close-up camera by pushing its camera-to-face distance $d_0$ to a small value $d_\text{init}$. At the same time, we adjust the focal length to maintain the eye position using the reparameterization method that will be described in \eqnref{alpha}.

\subsection{Optimization}
\paragraph{Optimization scheduling} 
When camera parameters are incorrect, 
the face latent code often overfits the target face, resulting in wrong geometry.
Therefore, we propose optimization scheduling, which sequentially optimizes the camera parameters, face latent code, and generator.

\paragraph{Focal length reparameterization}
We observe that the focal length is more sensitive than the camera-to-face distance in optimization (see \figref{dist_opt}). 
Therefore, we propose to relate the focal length to the camera-to-subject distance to limit the degree of freedom. 

Suppose the world-to-camera transformation is:
\begin{equation}
    \left[ \begin{array}{c}
    \mathbf{p}_c \\
    1\\
\end{array} \right] =\left[ \begin{matrix}
    \mathbf{R} &    \mathbf{t}\\
    0 & 1\\
\end{matrix} \right] \left[ \begin{array}{c}
    \mathbf{p}_w\\
    1\\
\end{array} \right] \,,
\end{equation}
where $\mathbf{R} = [\mathbf{r}_x, \mathbf{r}_y, \mathbf{r}_z]^T \in \mathds{R}^{3\times3}$ is the rotation matrix and $\mathbf{t}=[t_x, t_y, t_z]^T\in \mathds{R}^{3\times1}$ is the translation vector.
The intrinsic matrix $\mathbf{K}$ transforms a point from camera space to the image plane as:
\begin{equation}
z_c    \left[ \begin{array}{c}
    u\\
    v\\
    1\\
\end{array} \right] =  \mathbf{K} \, \mathbf{p}_c = \left[ \begin{matrix}
    f&      0&      c_x\\
    0&      f&      c_y\\
    0&      0&      1\\
\end{matrix} \right] \mathbf{p}_c \,.
\end{equation}
When adjusting the translation $t_z$, we relate the focal length $f$ to $t_z$ by ensuring the eye position remains unchanged. The relation is given by:
\begin{equation}\label{eq:alpha}
   f = \alpha f_0\,,\, \, \, \, \text{where}\, \, \alpha = (d_0 - (t_{z0} - t_z)) / d_0 \,,
\end{equation}
$d_0$ represents the initialization of camera-to-eye distance. The derivation can be found in the Appendix.
During optimization, we update the
intrinsic matrix by 
\begin{equation}\label{eq:intrinsic}
\mathbf{K} =
\begin{bmatrix}
\gamma\alpha f_0 & 0 & c_x\\
0 & \gamma\alpha f_0 & c_y\\
0 & 0 & 1
\end{bmatrix},
\end{equation}
where $\gamma$ is a learnable parameter with a small learning rate to accommodate error resulting from approximation.

\paragraph{Rotation reparameterization}
{Besides focal length parameterization, we also reparametrize the rotation matrix $\mathbf{R}$ 
to ensure orthogonality and reduce the degree of freedom:
\begin{equation}
\mathbf{R} =
\begin{bmatrix}
| & | & | \\
\bb_x & \bb_y & \bb_z \\
| & | & |
\end{bmatrix} 
= F\left( \mathbf{Q} \right) = 
F\left(
\begin{bmatrix}
| & | \\
\ba_1 & \ba_2 \\
| & |
\end{bmatrix}\right),
\label{eq:rotation}
\end{equation}
where $\bb_x, \bb_y, \bb_z \in \mathds{R}^3$ are $\bb_x = N(\ba_1)$, $\bb_y = N(\ba_2 - (\bb_x \cdot \ba_2) \bb_x)$, and $\bb_z = \bb_x \times \bb_y$, and $N(\cdot)$ denotes $L2$ norm.}

\subsection{Loss functions}
\paragraph{Landmark regularization}
The photometric loss function used in GAN inversion is ineffective for representing perspective changes. 
Therefore, we use an additional landmark loss to increase the sensibility of camera-to-subject variation.
We use the dense landmarks estimated from MediaPipe~\cite{lugaresi2019mediapipe} and calculate their $L2$ distances. Since there exist many unreliable landmarks, such as the occluded regions, we define an uncertainty-based landmark loss:
\begin{equation}\label{eq:landmark-loss}
    \mathcal{L}_{\text{landmark}}(m) = \sum_{i=1}^{\lVert\mathcal{M}\rVert} \left( \log\left(\sigma_i^2\right) + \frac{\lVert m_i - m^{\prime}_i\rVert_2^2}{2\sigma^2_i} \right),
\end{equation}
where $m\in \mathcal{M}$ is the normalized 
3D coordinates of the landmarks and $\lVert\mathcal{M}\rVert$ equals 468.
$\sigma$ is a learnable parameter to control the uncertainty.

\paragraph{Masked loss}
Close-up portraits often have faces that extend close to the image boundary, creating issues with the crop operation and potentially causing the cropped image to have an incomplete face and black boundaries. As a result, directly fitting such images may yield unusual facial features. To address this concern, we implement a masked loss, which allows us to ignore the out-of-boundary information. 

\subsection{Perspective-aware manipulation}
{
After 3D GAN inversion, we acquire optimized parameters to reconstruct the input face and manipulate camera settings to render virtual images $I_\text{novel}$. To correct face perspective distortion, we increase the camera-to-subject distance. We also adjust the focal length simultaneously to maintain a similar face size as the input according to \eqnref{alpha}.
}

\section{\revise{Extension for Full-frame Image}}

\revise{
Since face GANs can only process cropped face regions, 
to render a physically plausible full-frame image, we develop the geometry-aware stitching (\figref{method})
to extend the core distortion correction method to full-frame images.
}

\revise{
The basic idea is similar to STIT~\cite{tzaban2022stitch} that fine-tunes the generator with frozen inverted face latent code by minimizing the gap between the border pixels of the generated face and their corresponding pixels in the input image. As a result, the refined generator renders face images that can be seamlessly blended with the full image without visible inconsistencies.
}

\revise{
However, applying STIT~\cite{tzaban2022stitch} directly is infeasible. Because the perspective manipulation step yields a face image $I_\text{novel}$ with different camera parameters from the input full image $I_\text{real}$, leading to \emph{geometric inconsistencies} between them.
Merely fine-tuning the generator and then blending the generated face image and the input full image can reduce seams but introduce suspicious distortion, such as a disproportionately large face and a slim neck.
To overcome the challenge, our method reprojects the background with the camera parameters of the generated face, followed by the stitching tuning and blending steps.
}

\subsection{Reprojection}

We can effectively mitigate geometric misalignment issues by reprojecting the input image using the same camera parameters as the rendered face (shown in \figref{method}a). This reprojection process relies on point clouds. Initially, we acquire the depth map $\mathbf{d}_\text{near}^\text{full}$ for the input image through a monocular depth estimator~\cite{ranftl2020towards}. However, direct utilization is impossible since the depth map's scale differs from the rendered face's. Maintaining aligned depth maps for the entire image and the rendered face image becomes crucial.

To achieve this, we render the depth map $\mathbf{d}_\text{near}^\text{render}$ for the cropped face using the 3D GAN and align the monocular depth with it. This alignment is accomplished by minimizing the least square error:
\begin{equation}\label{eqn:align}
    \argmin_{s,b}\sum\lVert\left(s \times \texttt{Crop}({\mathbf{d}_\text{near}^\text{full}} \odot \Psi) + b\right) - \mathbf{d}_\text{near}^\text{render}\rVert_2^2\,,
\end{equation}
where $s$ and $b$ are the scale and shift parameters, $\odot$ is the element-wise multiplication, and $\Psi$ masks non-face regions.
But the aligned depth  $\mathbf{d}_\text{near}^\text{align} = s \odot \mathbf{d}_\text{near}^\text{full} + b$
is still diverse from the rendered face depth due to the limitation of the monocular depth estimator.
To refine it, we use the rendered face depth for the face region and   
use Poisson blending~\cite{perez2003poisson} to propagate the face depth to surrounding regions, \eg, body, hair. 
The content condition is based on the rendered face depth, while the gradient follows the monocular depth.
As the propagation proceeds from inner to outer regions, we set an outer boundary  $\mathbf{d}_\text{near}^\text{border}$ using the aligned depth map as the constraint.

Following propagation, we obtain $\mathbf{d}_\text{near}^\text{blend}$, a fine-grained depth map aligning with the rendered face depth. We then project the entire image to a longer distance using 3D GANs' camera parameters and the refined depth map.

\subsection{Stitch tuning}
Given the reprojected full image $I_\text{real}^\text{warp}$, we follow \cite{tzaban2022stitch} to fine-tune the generator's weights $\vartheta$ (as depicted in \figref{method}b).
We use a border loss to achieve a closely-matched border between our refined face image $I_\text{novel}^\text{refine}$ and the warped full image:
\begin{equation}
   \mathcal{L}_\text{border} = \lVert I_\text{novel}^\text{refine}\odot \tilde{\Psi} - \texttt{Crop}(I_\text{real}^\text{warp}) \odot \tilde{\Psi} \rVert_2^2\,, 
\end{equation}
where $\tilde{\Psi}$ is the border mask.
Likewise, we maintain the integrity of the content in our synthesis via a content loss:
\begin{equation}
   \mathcal{L}_\text{content} = \lVert I_\text{novel}^\text{refine}\odot \hat{\Psi} - I_\text{novel}\odot \hat{\Psi} \rVert_2^2\,, 
\end{equation}
where $\hat{\Psi}$ denotes the face inner region mask.

\subsection{Blending}
Finally, we blend the refined synthetic face image and the warped full image to produce an entire image virtually captured at a long distance, as shown in \figref{method}c. 
Note that if the inverted face loses details, we 
can alleviate such artifacts by warping the residual between input and inversion using the rendered depth map, then add it to the final images.

\section{Experiments}
\label{sec:exp}

\newlength{\quantimgw}
\setlength{\quantimgw}{0.24\linewidth}
\newcommand{\quantimgrow}[1]{            
	\fbox{\includegraphics[width=\quantimgw]{quant/#1_2.jpg}} &
	\fbox{\includegraphics[width=\quantimgw]{quant/#1_2_fried.jpg}} &
        \fbox{\includegraphics[width=\quantimgw]{quant/#1_2_shih.jpg}} &
        \fbox{\includegraphics[width=\quantimgw]{quant/#1_2_3dpx.jpg}} &
        \fbox{\includegraphics[width=\quantimgw]{quant/#1_2_hfgi3d.jpg}} &
        \fbox{\includegraphics[width=\quantimgw]{quant/#1_2_ret.jpg}} &
	\fbox{\includegraphics[width=\quantimgw]{quant/#1_2_ours.jpg}} &
	\fbox{\includegraphics[width=\quantimgw]{quant/#1_2_ref16.jpg}} 
}

\begin{figure*}
    \centering
    \footnotesize
     \begin{tabular}
    {   
        @{\hspace{0mm}}c@{\hspace{0.8mm}} 
        @{\hspace{0mm}}c@{\hspace{0.8mm}}
        @{\hspace{0mm}}c@{\hspace{0.8mm}} 
        @{\hspace{0mm}}c@{\hspace{0.8mm}} 
        @{\hspace{0mm}}c@{\hspace{0.8mm}} 
        @{\hspace{0mm}}c@{\hspace{0.8mm}} 
        @{\hspace{0mm}}c@{\hspace{0.8mm}}
        @{\hspace{0mm}}c@{\hspace{0.8mm}} 
    }
        \quantimgrow{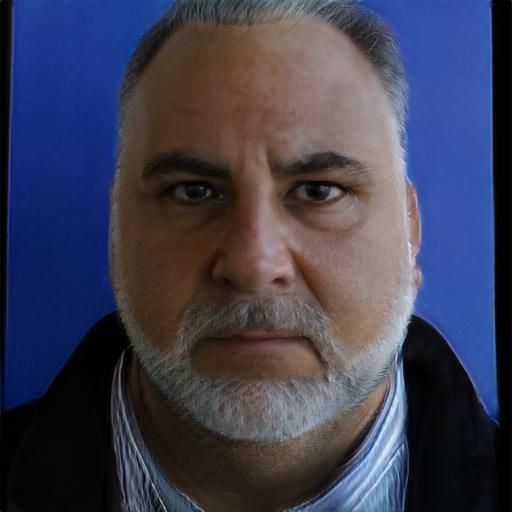}\\
        \quantimgrow{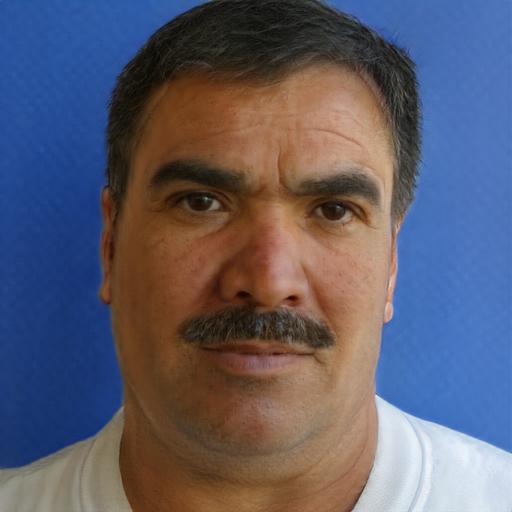}\\
        \quantimgrow{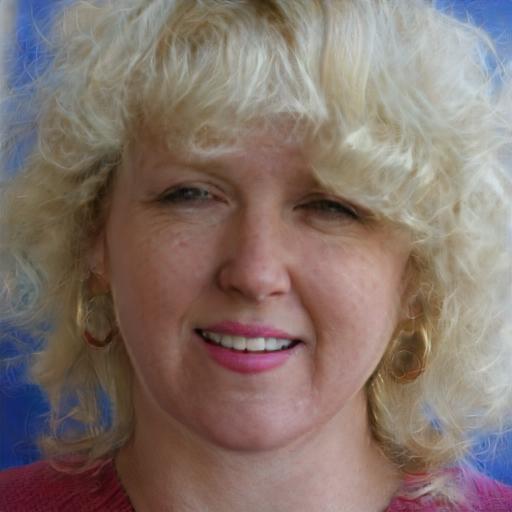}\\
        \setarial{Input} & Fried's~\cite{fried2016perspective}  & \revise{Shih's~\cite{shih2019distortion}} & \revise{3DP~\cite{Shih3DP20}} & \revise{HFGI3D~\cite{xie2023high}} & \revise{Triplanenet~\cite{bhattarai2024triplanenet}} & \setarial{Ours} & \setarial{Reference}
    \end{tabular}
    \caption{\tb{Qualitative comparisons on the CMDP dataset~\cite{burgos2014distance}.}
    Results of \cite{fried2016perspective} are from their website. Our method renders faces closer to their references while preserving the identity. 
    }
    \label{fig:quant-img}
\end{figure*}

\subsection{Experimental setup}

\paragraph{Dataset}
We use three different datasets for evaluation:
\begin{itemize}
    \item \textbf{Caltech Multi-Distance Portraits (CMDP) \cite{fried2016perspective}}: This dataset contains portrait images of different people taken from various distances. 
It provides the same identities taken from different distances. We use the CMDP dataset for quantitative evaluations. 

    \item \textbf{USC perspective portrait database~\cite{Zhao-2019-ICCV}}: This database contains images with single faces with different levels of perspective distortions. 
    There are no references or ground truth images, so we only use these images for visual comparisons.

    \item \textbf{In-the-wild images}: We also collect many in-the-wild photos online with severe perspective distortions on faces. 
    We use these images for visual comparisons.
\end{itemize}

\newlength{\dollyw}
\setlength{\dollyw}{0.24\linewidth}
\newcommand{\dollyrow}[1]{            
	\fbox{\includegraphics[width=\dollyw]{comparison/#1_input.jpg}} &
	\fbox{\includegraphics[width=\dollyw]{comparison/#1_sig16.jpg}} &
        \fbox{\includegraphics[width=\dollyw]{comparison/#1_iccv19.jpg}} &
        \fbox{\includegraphics[width=\dollyw]{qualitative/#1_shih.jpg}} &
        \fbox{\includegraphics[width=\dollyw]{qualitative/#1_3dpx.jpg}} &
        \fbox{\includegraphics[width=\dollyw]{qualitative/#1_original_pti+chen_16_.jpg}} &
	\fbox{\includegraphics[width=\dollyw]{qualitative/#1_triplane.jpg}} &
	\fbox{\includegraphics[width=\dollyw]{comparison/#1_ours_transferred_.jpg}} 
}

\newcommand{\dollyrowtrim}[5]{            
	\fbox{\adjincludegraphics[trim={#2 {#3} {#4} {#5}}, clip, width=\dollyw]{comparison/#1_input.jpg}} &
        \fbox{\adjincludegraphics[trim={#2 {#3} {#4} {#5}}, clip, width=\dollyw]{comparison/#1_sig16.jpg}} &
        \fbox{\adjincludegraphics[trim={#2 {#3} {#4} {#5}}, clip, width=\dollyw]{comparison/#1_iccv19.jpg}} &
        \fbox{\adjincludegraphics[trim={#2 {#3} {#4} {#5}}, clip, width=\dollyw]{qualitative/#1_shih.jpg}} &
        \fbox{\adjincludegraphics[trim={#2 {#3} {#2} {#5}}, clip, width=\dollyw]{qualitative/#1_3dpx.jpg}} &
        \fbox{\adjincludegraphics[trim={#2 {#3} {#4} {#5}}, clip, width=\dollyw]{qualitative/#1_original_pti+chen_16_.jpg}} &
        \fbox{\adjincludegraphics[trim={#2 {#3} {#4} {#5}}, clip, width=\dollyw]{qualitative/#1_triplane.jpg}} &
	\fbox{\adjincludegraphics[trim={#2 {#3} {#4} {#5}}, clip, width=\dollyw]{comparison/#1_ours_transferred_.jpg}}
}

\begin{figure*}
    \centering
    \footnotesize
     \begin{tabular}
    {   
        @{\hspace{0mm}}c@{\hspace{0.8mm}} 
        @{\hspace{0mm}}c@{\hspace{0.8mm}}
        @{\hspace{0mm}}c@{\hspace{0.8mm}} 
        @{\hspace{0mm}}c@{\hspace{0.8mm}} 
        @{\hspace{0mm}}c@{\hspace{0.8mm}} 
        @{\hspace{0mm}}c@{\hspace{0.8mm}}
        @{\hspace{0mm}}c@{\hspace{0.8mm}} 
        @{\hspace{0mm}}c@{\hspace{0.8mm}}
    }
        \dollyrowtrim{x14-FJHAPL}{0.1\width}{0}{0.1\width}{0}\\
        \dollyrow{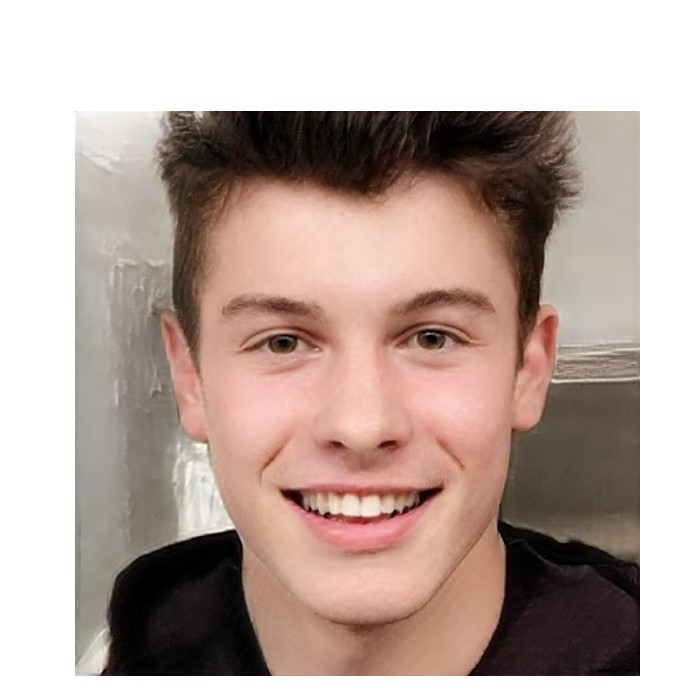}\\
         \dollyrow{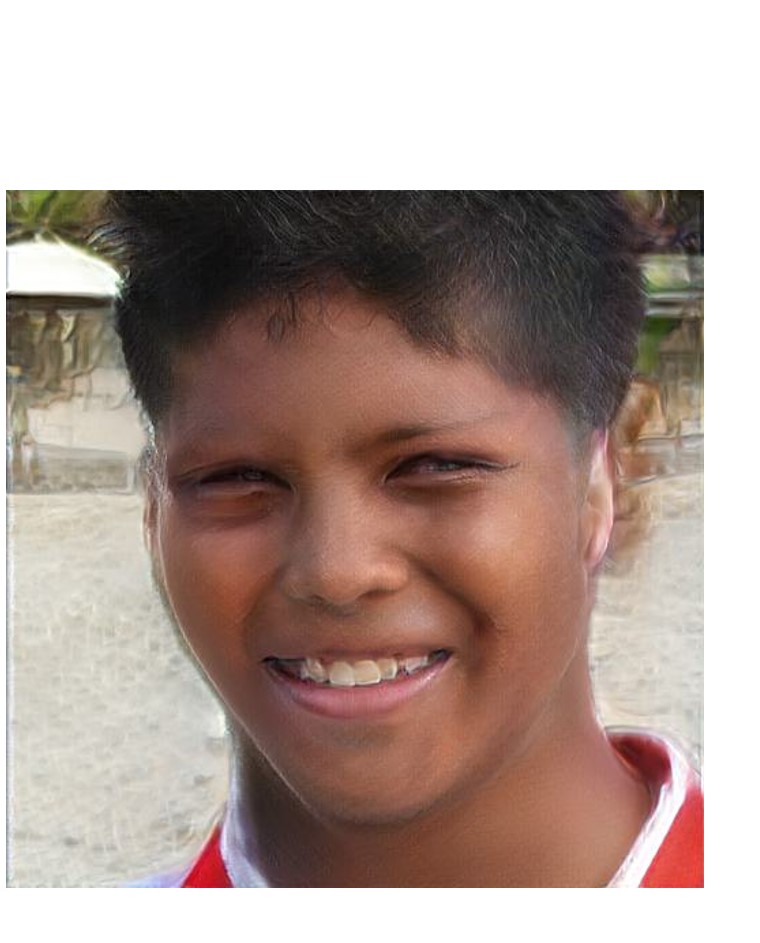}\\
        \setarial{Input} &   Fried's~\setarial{\cite{fried2016perspective}} & Zhao's~\setarial{\cite{Zhao-2019-ICCV}} & \revise{Shih's~\cite{shih2019distortion}} & \revise{3DP~\cite{Shih3DP20}} & \revise{HFGI3D~\cite{xie2023high}} & \revise{Triplanenet~\cite{bhattarai2024triplanenet}} & \setarial{Ours}
    \end{tabular}
    \caption{\textbf{Qualitative comparisons on images collected by \cite{Zhao-2019-ICCV}.} Results of compared methods \cite{fried2016perspective,Zhao-2019-ICCV} are from \cite{Zhao-2019-ICCV}. 
    Our method produces the least distorted and the most natural perspective correction results. Note that with the help of 3D GAN, our method can generate the ear that originally occluded in the input images.
    }
    \label{fig:full-img}
\end{figure*}

\paragraph{Compared methods}
We compare our method with: 
\begin{itemize}
    \item \textbf{Portrait perspective undistortion:} Fried's~\cite{fried2016perspective} and Zhao's~\cite{Zhao-2019-ICCV} focus on the same task as us but they are 2D warping-based solutions. Since neither releases official implementations, we re-implement the method of \cite{fried2016perspective}. In addition to comparing with our own implementation of the two methods, we also obtained several results from the website of \cite{fried2016perspective} and the authors of \cite{Zhao-2019-ICCV} for comparison.
    \item \revise{\textbf{Wide-angle undistortion methods:} Shih's~\cite{shih2019distortion} is a technique that solves a different undistortion problem with us: distortion caused by a wide-angle lens. Their basic idea is to apply the stereographic projection to the distorted image. 
    }
    \item \textbf{2D/3D GAN inversion methods:} PTI~\cite{Roich-2021-TOG-PTI}, Ko's~\cite{ko20233d}, \revise{HFGI3D~\cite{xie2023high}, and Triplanenet~\cite{bhattarai2024triplanenet}}. Although not explicitly dealing with portrait perspective correction, 
    these 2D/3D GAN inversion methods enable 3D GANs to generate novel views from a single image.
    \item \revise{\textbf{3D photography:} 3DP~\cite{Shih3DP20} is a method that can render novel views from a single RGB-D image. 
    }
\end{itemize}

\begin{table}[!t]
\centering
\caption{
\tb{Quantitative comparison on the CMDP dataset~\cite{burgos2014distance}}. We evaluate 43 faces projected from 60 cm to 480~cm. The photometric loss is low because reference images are captured asynchronously with different camera parameters from the inputs, resulting in different
appearances and poses. ‘W’ represents warping-based and `G' denotes GAN inversion-based. $^\star$Results from the official \href{https://www.ohadf.com/projects/perspective-portraits/results/view_results.html}{website}.
         $^\dagger$Our re-implementation. Although the results differ from the original ones, the metric scores are comparable. 
}
\label{tab:quant}
\setlength{\tabcolsep}{3pt}
	
        \resizebox{\columnwidth}{!}
        {
	\begin{tabular}{rcccccc}
            \toprule
		{Method} &   Type &  LMK-E$\downarrow$ & PSNR$\uparrow$ & SSIM$\uparrow$ & LPIPS$\downarrow$ & ~~ID$\uparrow$~~~\\
		\cmidrule(lr){1-1}\cmidrule(lr){2-2}\cmidrule(lr){3-7}
            $^\star$Fried's~\cite{fried2016perspective} & W  & \cellcolor{third}0.175 & 15.41 & \cellcolor{second}0.724 & \cellcolor{second}0.188 & \cellcolor{best}0.893 \\
		  $^\dagger$Fried's~\cite{fried2016perspective}  & W & \cellcolor{second}0.165 & 14.41 & 0.716 & 0.208 & \cellcolor{second}0.860 \\
            \revise{Shih's~\cite{shih2019distortion}} & W  & 0.236  & 12.95  & 0.696  & 0.258  &  0.855\\
            \revise{3DP~\cite{Shih3DP20}} & W  & 0.195  & 13.08  & 0.696  & 0.268  & 0.847\\
            PTI~\cite{Roich-2021-TOG-PTI} & G  & 0.191 & \cellcolor{second}15.92 & 0.717 & \cellcolor{third}0.197  & 0.758     \\
           Ko's~\cite{ko20233d}  & G &  0.180 & 15.41 & 0.710 & 0.206  & 0.689   \\
            \revise{HFGI3D~\cite{xie2023high}}  & G & 0.177 & \cellcolor{third}15.75 & \cellcolor{second}0.724 & 0.198 & 0.829 \\
            \revise{Triplanenet~\cite{bhattarai2024triplanenet}} & G   & 0.188  & 14.80  & 0.705  & 0.243 & 0.812\\
            Ours & G &  \cellcolor{best}0.138  &  \cellcolor{best}17.52  &  \cellcolor{best}0.747  &  \cellcolor{best}0.167 & \cellcolor{third}0.859 \\
		\bottomrule
	\end{tabular}
        }
\end{table}

\newlength{\qualfw}
\setlength{\qualfw}{0.24\linewidth}
\newcommand{\qualfrow}[1]{            
	\includegraphics[width=\qualfw]{qualitative/#1_input.jpg} &
 
	\includegraphics[width=\qualfw]{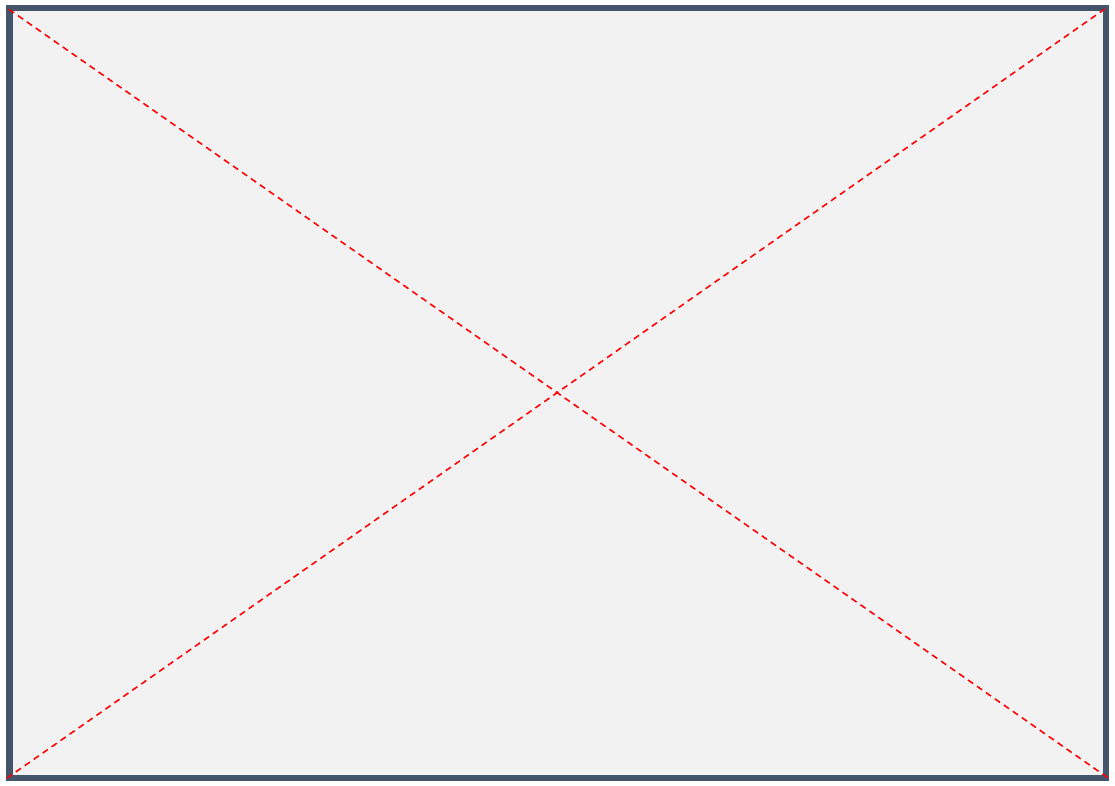}  &
 \includegraphics[width=\qualfw]{figure/figuresample.pdf}  &
	\includegraphics[width=\qualfw]{qualitative/#1_oursx4.jpg} &
	\includegraphics[width=\qualfw]{qualitative/#1_oursx8.jpg} 
}
\newcommand{\qualfrowx}[5]{            
\fbox{\adjincludegraphics[trim={{#2\width} {#3\height} {#4\width} {#5\height}},clip,width=\qualfw]{qualitative/#1_input.jpg}} & 
\fbox{\adjincludegraphics[trim={{#2\width} {#3\height} {#4\width} {#5\height}},clip,width=\qualfw]{qualitative/#1_sig16x8.jpg}} & 
\fbox{\adjincludegraphics[trim={{#2\width} {#3\height} {#4\width} {#5\height}},clip,width=\qualfw]{qualitative/#1_wacvx8.jpg}} & 
\fbox{\adjincludegraphics[trim={{#2\width} {#3\height} {#4\width} {#5\height}},clip,width=\qualfw]{qualitative/#1_ours_transferred.jpg}} & %
}
\providelength\liftup
\setlength\liftup{-3.5mm}
\begin{figure*}
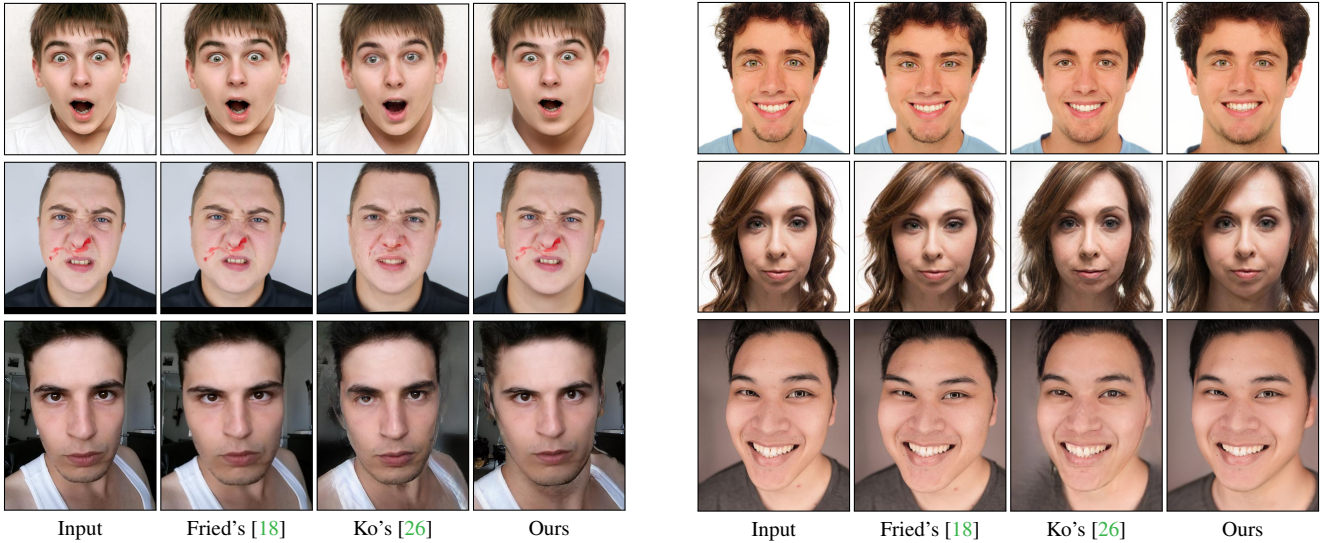

    \centering
    \footnotesize
    \begin{minipage}[c]{0.472\textwidth}
     \begin{tabular}
    {   
        @{\hspace{0mm}}c@{\hspace{0.5mm}} 
        @{\hspace{0mm}}c@{\hspace{0.5mm}}
        @{\hspace{0mm}}c@{\hspace{0.5mm}}
        @{\hspace{0mm}}c@{\hspace{0.5mm}}
        @{\hspace{0mm}}c@{\hspace{0.5mm}}
    }
    \qualfrowx{tp3-CTTSKA}{0}{0}{0}{0}\\
    \qualfrowx{x22-DVPAXP}{0}{0}{0}{0}\\
    \qualfrowx{x3-VHQTSI}{0.1}{0}{0.1}{0}\\
    Input & Fried's~\cite{fried2016perspective} &  Ko's~\cite{ko20233d} & Ours \\
    \end{tabular}
    \end{minipage}
    \hfill
    \begin{minipage}[c]{0.472\textwidth}
     \begin{tabular}
    {   
        @{\hspace{0mm}}c@{\hspace{0.5mm}} 
        @{\hspace{0mm}}c@{\hspace{0.5mm}}
        @{\hspace{0mm}}c@{\hspace{0.5mm}}
        @{\hspace{0mm}}c@{\hspace{0.5mm}}
        @{\hspace{0mm}}c@{\hspace{0.5mm}}
    }
    \qualfrowx{ct2-CGSVRN}{0}{0}{0.0}{0}\\
    \qualfrowx{tp1-PAFAEH}{0}{0.054}{0.05}{0}\\
    \qualfrowx{x26-FVPOQG}{0.1}{0}{0.15}{0.05}\\
    Input & Fried's~\cite{fried2016perspective} &  Ko's~\cite{ko20233d} & Ours \\
    \end{tabular}
    \end{minipage}
    \caption{\textbf{Visual results for our collected severely distorted in-the-wild face images.} 
    We enlarge the camera-to-subject distance to 
    $\times8$ times the estimated distance. 
    Our method performs well in dealing with these seriously distorted faces and recovering occluded regions, such as ears. 
    }
    \label{fig:qualitative}
\end{figure*}

\paragraph{Evaluation metrics}
We use five evaluation metrics to evaluate the performance of portrait perspective correction:
\begin{itemize}
\item  \textbf{Euclidean distance landmark error}: 
We first align all output faces, and their corresponding reference faces according to the dense facial landmarks detected via mediapipe~\cite{lugaresi2019mediapipe}. 
We follow a similar alignment method by StyleGAN~\cite{karras2019style} to align the landmarks.
We then calculate the normalized landmark distance error in the 2D Euclidean space. 

\item \textbf{Photometric errors PSNR, SSIM, and LPIPS}: 
We also calculate photometric errors between the aligned output images and corresponding references, including PSNR, SSIM~\cite{wang2004image}, and LPIPS~\cite{zhang2018unreasonable}.
We use a tri-map free matting algorithm~\cite{ke2022MODNet} to remove the background and calculate the photometric distances on the masked foreground. 

\item \textbf{Identity similarity}: {We use ArcFace~\cite{deng2019arcface} to extract features for the masked face foregrounds and compute the cosine distance between facial features of output images and reference images. }
\end{itemize}

\subsection{Quantitative evaluation}
We evaluate our method on the CMDP dataset~\cite{burgos2014distance}, and the results in \tabref{quant} indicate:
(1) Our method outperforms others in most metrics with a large margin;
(2) All methods, including ours, exhibit inferior performance in identity preservation compared to the original version of \cite{fried2016perspective}. This is primarily due to the significance of face details in calculating identity metrics.  The original version of \cite{fried2016perspective} has subtle manipulations and retains many details. 
GAN inversion-based methods have the lowest identity score among all methods because they may lose some crucial details.
(3) Despite the limitations of GAN inversion, our method achieves comparable results to our reimplementation of the warping-based method \cite{fried2016perspective} in the identity metric.

\subsection{Qualitative evaluation}

{
We evaluate our proposed method on cropped face images used by previous methods, and the comparisons are presented in \figref{quant-img} and \figref{full-img}.
The changes to distorted faces introduced by \cite{fried2016perspective} and \cite{shih2019distortion} are infinitesimal. In contrast, evident changes can be observed when distorted faces are corrected by \cite{Zhao-2019-ICCV} and 3DP~\cite{Shih3DP20}. However, their corrections lead to amplified distortions, where the middle part of faces is less distorted, but the head and chin shapes still appear peculiar (\figref{full-img}). 
Our method generates faces with fewer perspective distortions while preserving identity. Moreover, with the aid of 3D GAN, our approach can generate occluded parts present in the original input images, such as ears.
\revise{
It is worth noting that other GAN inversion-based solutions~\cite{xie2023high,bhattarai2024triplanenet} struggle to recover the correct face shape.}

We further demonstrate this advantage on our collected \textbf{in-the-wild} faces with severe distortions and showcase the perspective distortion correction results in \figref{qualitative}. We notice that the re-implemented method~\cite{fried2016perspective} performs similarly to \cite{Zhao-2019-ICCV}. Additionally, we observe that the GAN inversion-based method~\cite{ko20233d} encounters local minima and generates faces with incorrect shapes. The visual results clearly demonstrate that our perspective-aware 3D GAN inversion proves to be an effective approach for portrait perspective correction, outperforming the warping-based method~\cite{fried2016perspective} and the existing 3D GAN inversion-based method~\cite{ko20233d}.
}

\subsection{Full-image qualitative evaluation}

We validate our system's ability to process \textbf{in-the-wild full} images, as demonstrated by the visually pleasing results in \figref{teaser} and \figref{full-frame}. In comparison, the other methods fail to reduce perspective distortion or generate harmonious results effectively.
Specifically, (1) the changes caused by Fried's~\cite{fried2016perspective} are subtle, and the manipulated face remains distorted. (2) Zhao's~\cite{Zhao-2019-ICCV} significantly alters the face, but the result still exhibits an asymmetric face shape, weird head and chin shapes, and inconsistency between the body and face. (3) Although 3DP~\cite{Shih3DP20} can manipulate the body and somewhat mitigate face distortion by using the depth from 3D GAN, the face is still distorted. (4) Combining Ko's~\cite{ko20233d} and STIT~\cite{tzaban2022stitch} results in a seamless image but lacks harmony. On the other hand, our manipulated faces exhibit harmonious integration with corresponding bodies, with fewer distortions.

\subsection{Video evaluation}

In comparing our method with others in rendering dolly-zoom videos from distorted input, the results in  \emph{\textcolor{nice-red}{supplemental materials}} demonstrate that only our approach can consistently generate continuous dolly-zoom videos. In contrast, other methods show the following limitations:
(1) Fried's~\cite{fried2016perspective} corrects distortion but performs worse than ours, with minimal manipulation in non-face regions.
(2) 3DP~\cite{Shih3DP20} is unable to manipulate the face.
(3) Combining Ko's~\cite{ko20233d} with STIT~\cite{tzaban2022stitch} leads to serious distortion.

\newlength{\fullw}
\setlength{\fullw}{0.28\textwidth}
\newcommand{\fullrow}[1]{            
	\fbox{\includegraphics[width=\fullw]{comparison_full/#1_input.jpg}} &
	\fbox{\includegraphics[width=\fullw]{comparison_full/#1_sig16.jpg}} &
 \fbox{\includegraphics[width=\fullw]{comparison_full/#1_iccv19.jpg}}\\
 \setarial{Input} & Fried's~\setarial{\cite{fried2016perspective}} & Zhao's~\setarial{\cite{Zhao-2019-ICCV}}  \\
 	\fbox{\includegraphics[width=\fullw]{comparison_full/#1_3dp.jpg}} & 
        \fbox{\includegraphics[width=\fullw]{comparison_full/#1_stit.jpg}} &
       \fbox{\includegraphics[width=\fullw]{comparison_full/#1_ours.jpg}} \\
       \setarial{3DP~\cite{Shih3DP20}} + Rendered face depth & Ko's~\setarial{\cite{ko20233d}+STIT~\cite{tzaban2022stitch}} & \setarial{Ours} \\
}

\begin{figure*}
    \centering
    \footnotesize
     \begin{tabular}
    {   
        @{\hspace{0mm}}c@{\hspace{1mm}} 
        @{\hspace{0mm}}c@{\hspace{1mm}}
        @{\hspace{0mm}}c@{\hspace{1mm}}
    }
        \fullrow{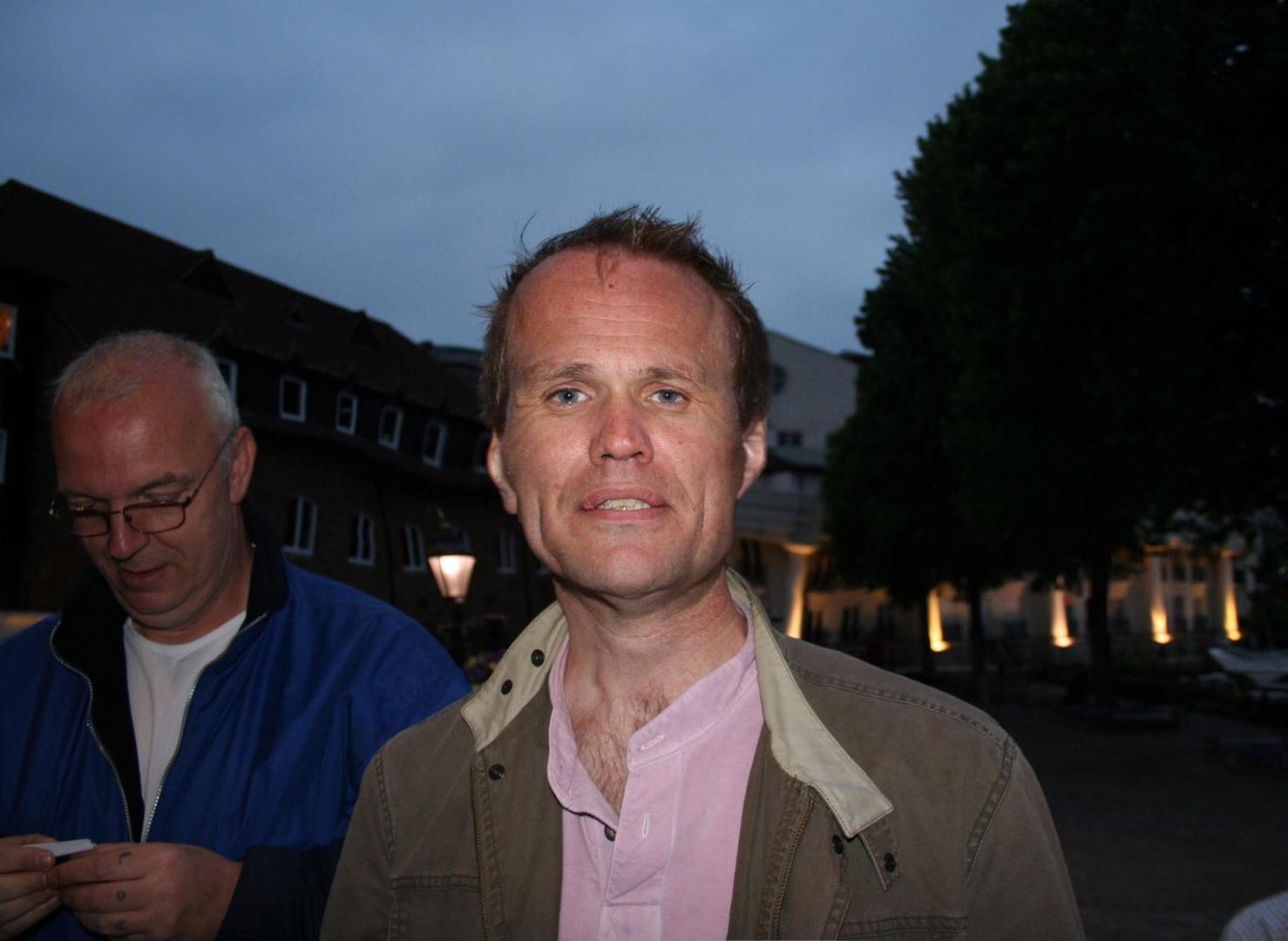}\\
    \end{tabular}
    \vspace{-4mm}
    \caption{\tb{Comparison on in-the-wild full images.} Results of compared methods~\cite{fried2016perspective,Zhao-2019-ICCV} are from \cite{Zhao-2019-ICCV}. Our system produces a visually pleasing result with the least distortions. Note that our rendered face is harmonious with the body. 
    }
    \label{fig:full-frame}
\end{figure*}

\begin{table}[!t]
    \centering
    \caption{\textbf{Quantitative results of ablation study.} 
    Focal length reparameterization and distance initialization are crucial. Removing any of them (v3 and v5) significantly degrades performance. 
    Optimization scheduling is important to avoid sub-optimal results. 
    Discarding camera optimization yields the worst photometric metric. 
    Our method achieves the best performance.
    }
    \setlength{\tabcolsep}{4pt}
    \footnotesize
    \begin{tabular}{lccccccc}
    \toprule
         &  \rotatebox{90}{\makecell{cam opt}} & \rotatebox{90}{rot. repa.} & \rotatebox{90}{focal repa.} & \rotatebox{90}{\makecell{schedule}}
         & \rotatebox{90}{\makecell{closeup}} & 
         LMK-E$\downarrow$ & LPIPS$\downarrow$  \\\cmidrule(lr){1-1}\cmidrule(lr){2-6}\cmidrule(lr){7-8}
         low bound (input) & -- & -- & -- & -- & -- & \textcolor{light-gray}{0.227} & \textcolor{light-gray}{0.249} \\
         (v0): w/o all & \xmark & \xmark & \xmark & \xmark  & \xmark  & 0.190 & 0.198\\
         (v1): w/o cam. opt. &  \xmark  & --  & \cmark & -- & \cmark  & 0.159 & 0.204 \\
         \revise{(v2): w/o rot. repa.} & \cmark & \xmark & \cmark & \cmark & \cmark & 0.167 & 0.203\\
         (v3): w/o focal repa. & \cmark & \cmark & \xmark & \cmark & \cmark & 0.183 & 0.200 \\
         (v4): w/o opt. sche. & \cmark & \cmark & \cmark & \xmark & \cmark & 0.151 & 0.182 \\
         (v5): w/o closeup cam & \cmark & \cmark & \cmark & \cmark & \xmark  & 0.185 & 0.198\\
         \rowcolor{verylight-gray}Ours & \cmark & \cmark & \cmark & \cmark & \cmark   & \textbf{0.138} & \textbf{0.167}\\\bottomrule 
    \end{tabular}
    \label{tab:ablation_table}
\end{table}

\newlength{\ablw}
\setlength{\ablw}{0.3\columnwidth}
\begin{figure}
    \centering
    
    \footnotesize
     \begin{tabular}
    {   
        @{\hspace{0mm}}c@{\hspace{0.8mm}} 
        @{\hspace{0mm}}c@{\hspace{0.8mm}}
        @{\hspace{0mm}}c@{\hspace{0.8mm}}
        @{\hspace{0mm}}c@{\hspace{0.8mm}} 
    }
\fbox{\includegraphics[width=\ablw]{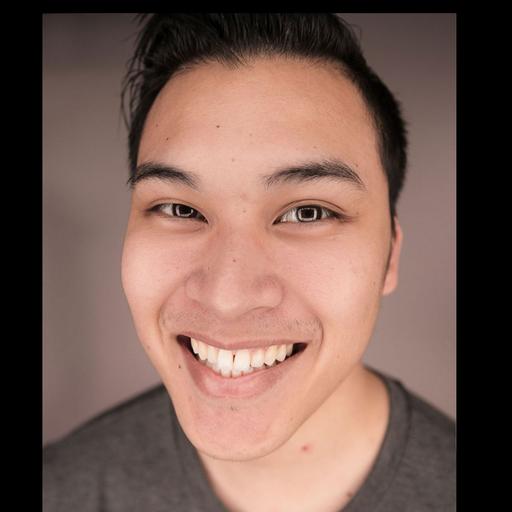}} &
\fbox{\includegraphics[width=\ablw]{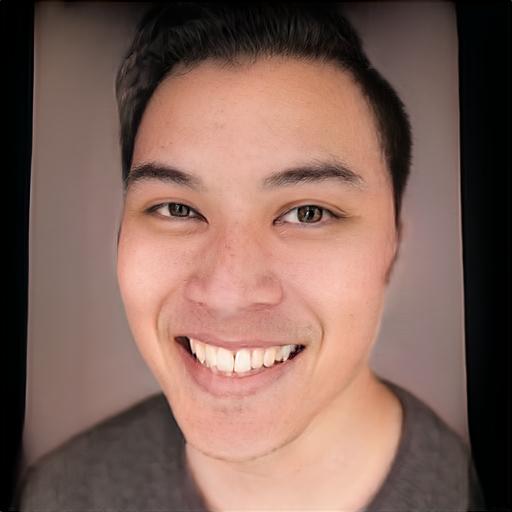}} &
\fbox{\includegraphics[width=\ablw]{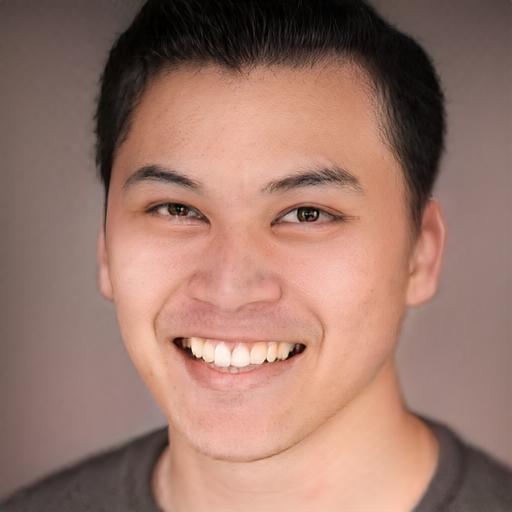}} \\
\setarial{Input} &
(v0) & 
(v1)  \\
\fbox{\includegraphics[width=\ablw]{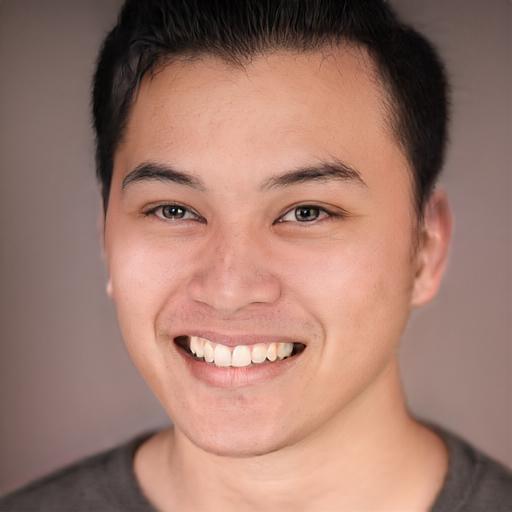}} &
\fbox{\includegraphics[width=\ablw]{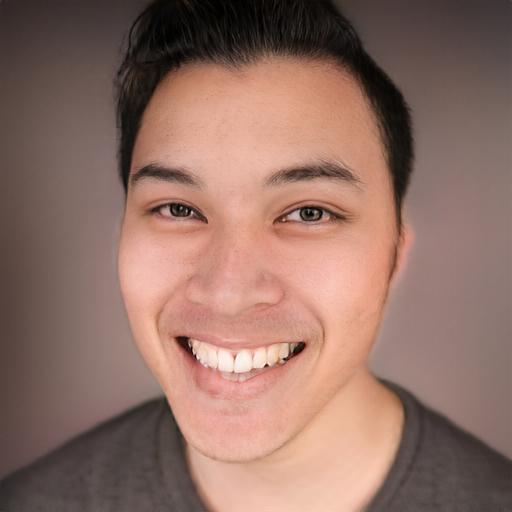}} &
\fbox{\includegraphics[width=\ablw]{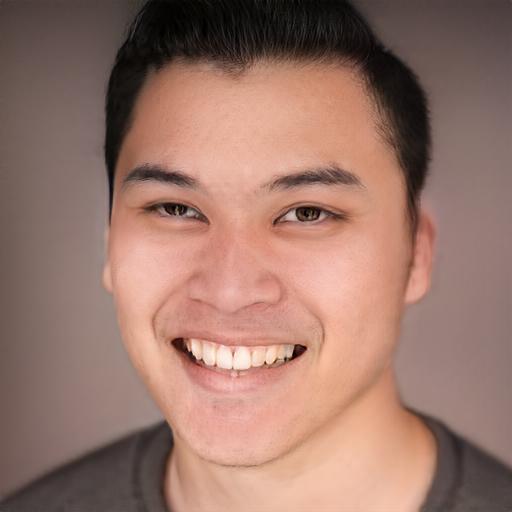}} \\
\setarial{(v2)} &
\setarial{(v3)} & 
\setarial{(v4)} \\
\fbox{\includegraphics[width=\ablw]{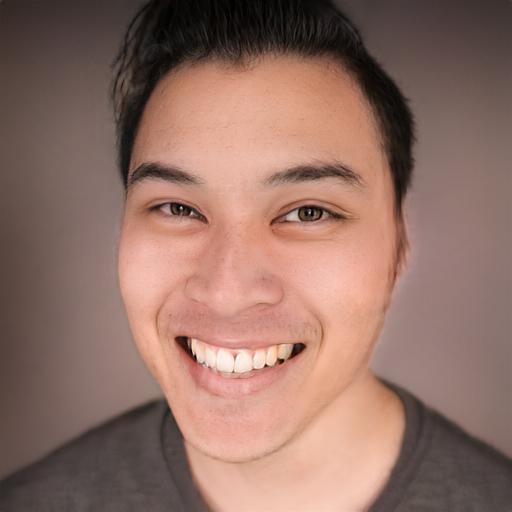}} &
\fbox{\includegraphics[width=\ablw]{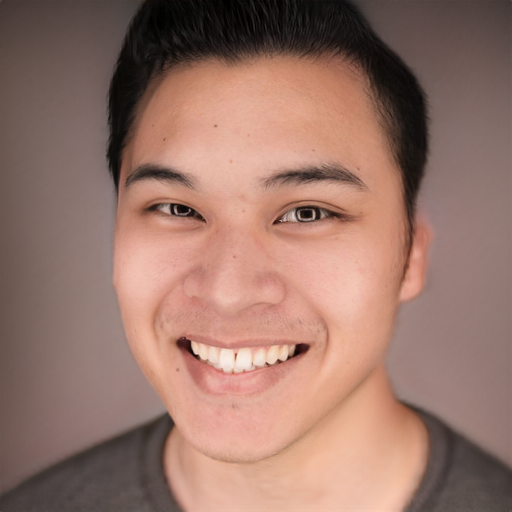}} &
\fbox{\includegraphics[width=\ablw]{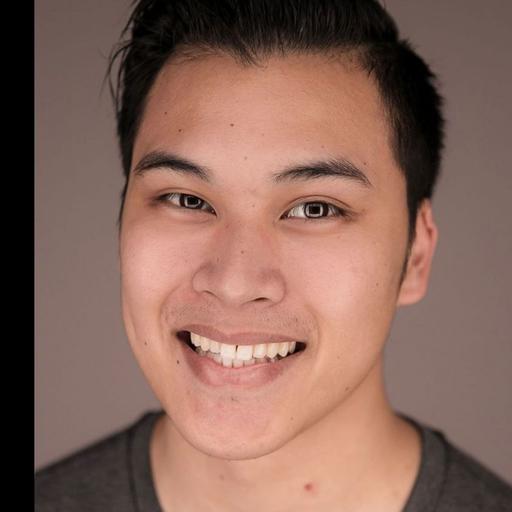}} \\
\setarial{(v5)}  & 
\setarial{Ours} & 
Reference\textsuperscript{\textcolor{red}{\textdagger}}\\
    \end{tabular}
    \vspace{-2mm}
    \caption{\textbf{Qualitative results of ablation study.}
    Our full model produces a visually pleasing result closest to the reference. 
    It cannot perform well if any of these designs are removed.
    Although quantitative results in \tabref{ablation_table} reveal that optimization scheduling is not dominant in our method, it is necessary to avoid sub-optimal results.
    \textsuperscript{\textcolor{red}{\textdagger}}Note that the reference is not the ground truth.
    }
    \label{fig:ablation-visual}
\end{figure}

\newlength{\qualw}
\setlength{\qualw}{0.28\textwidth}
\newcommand{\qualrow}[5]{            
	\fbox{\includegraphics[trim={#2 #3 #4 #5}, clip,width=\qualw]{dolly-zoom-out/#1_inputx.jpg}} &
        \fbox{\includegraphics[trim={#2 #3 #4 #5}, clip,width=\qualw]{ablation/#1_paste.jpg}} &
        \fbox{\includegraphics[trim={#2 #3 #4 #5}, clip,width=\qualw]{ablation/#1_paste_tune.jpg}}\\
        \setarial{Input} & \setarial{Direct paste} & \setarial{STIT~\cite{tzaban2022stitch}} \\
        \fbox{\includegraphics[trim={#2 #3 #4 #5}, clip,width=\qualw]{ablation/#1_3dpx.jpg}} &
        \fbox{\includegraphics[trim={#2 #3 #4 #5}, clip,width=\qualw]{ablation/#1_warp_paste.jpg}} &
	\fbox{\includegraphics[trim={#2 #3 #4 #5}, clip,width=\qualw]{dolly-zoom-out/#1_oursx8.jpg}}\\
    \setarial{Warp} & \setarial{Warp+Paste} &\setarial{Our final}\\
}
\begin{figure*}
    \centering
    \footnotesize
     \begin{tabular}
    {   
        @{\hspace{0mm}}c@{\hspace{1.2mm}} 
        @{\hspace{0mm}}c@{\hspace{1.2mm}}
        @{\hspace{0mm}}c@{\hspace{1.2mm}}
        @{\hspace{0mm}}c@{\hspace{1.2mm}} 
        @{\hspace{0mm}}c@{\hspace{1.2mm}} 
    }
    \qualrow{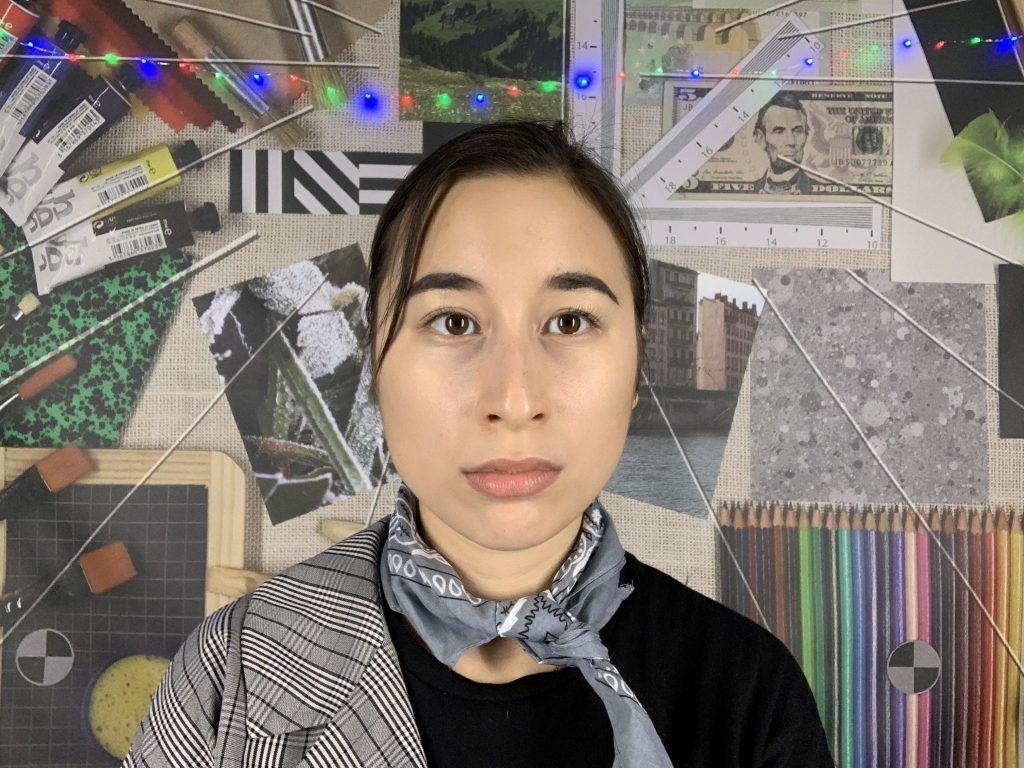}{0cm}{0cm}{0cm}{0cm}\\
    \end{tabular}
    \vspace{-2mm}
    \caption{\textbf{Qualitative results for ablation study of geometric-aware stitching.}  
    3D GANs can only reproject a cropped face image to a virtual far distance while leaving the rest of the image distorted. 
    Pasting the modified face back into the original image can lead to inconsistencies between the cropped face and the untouched regions.
    This geometry inconsistency cannot be reduced by the method \cite{tzaban2022stitch} used by 2D GAN inversion/manipulation.
    To address this issue, we reproject the background and fine-tune the generator to achieve seamless blending.
    }
    \label{fig:ablation-blend}
\end{figure*}

\subsection{Ablation study}
We conduct ablation studies on both the CMDP dataset and our collected seriously distorted face images. The results are presented in \tabref{ablation_table} and \figref{ablation-visual}. Without camera optimization or any of our proposed designs for easing optimization, the face parameter gets stuck in a sub-optimal solution, leading to poor performance. The proposed focal length reparameterization and distance initialization are crucial for achieving good results, and removing any of them results in a significant degradation in performance, with the reconstructed face geometry being wrong and the corrected image remaining distorted as the input. While removing optimization scheduling, \revise{rotation reparameterization} and camera optimization can still correct the distortion to some extent, it is more prone to fall into a local minimum, generating a face far away from the reference. \revise{The rotation reparameterization reduces the degree of freedom and regularizes the orthogonality of the rotation matrix.}

Our pipeline's ablation studies investigate the stitching post-processing, as shown in \figref{ablation-blend}. When we directly paste the manipulated face into the input image, it results in an inconsistency between the face and body parts. However, we can achieve seamless blending with further processing, producing a more harmonious and natural result.

\subsection{Manipulation to different distances}

\revise{We assess our model's ability to render images across various camera-to-subjective distances using the CMDP~\cite{burgos2014distance} dataset. This dataset comprises images of subjects captured from seven distinct distances. We select the closest image for each subject as our input and then project it into the remaining six distances. As shown in Figure \ref{fig:quant_dist}, our method consistently outperforms the baseline PTI~\cite{Roich-2021-TOG-PTI} across all distances, with its superiority increasing as the distance grows.}

\begin{figure}
    \centering 
    \includegraphics[width=0.88\columnwidth]{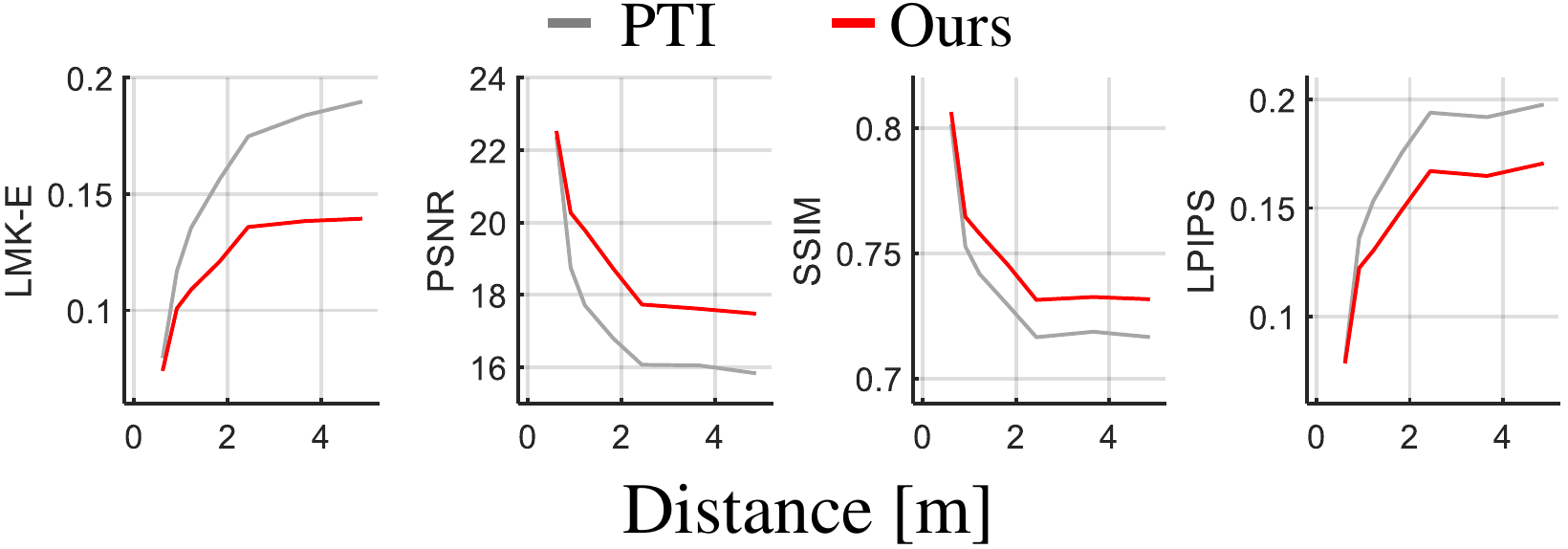}
    \caption{\revise{\tb{Evaluation of rendering at different distances.} We projected the input distorted images to various distances, with the result at each distance being an average of 43 faces. Notably, our method consistently outperforms PTI~\cite{Roich-2021-TOG-PTI} by a significant margin as the projected distance increases.
    }}
    \label{fig:quant_dist}
\end{figure}

\subsection{User study}
\revise{We conduct two user studies to compare our perspective 3D GAN inversion method with conventional GAN inversion method PTI~\cite{Roich-2021-TOG-PTI} with estimated cameras. In the first study, we presented results on 15 CMDP images alongside reference images to 56 participants and asked them to identify which method yields an image that closely resembles the reference. In the second study, we showed results on 10 in-the-wild images to 25 users and asked which method produces a less distorted image. Results in \figref{user} demonstrate that our method consistently outperforms PTI~\cite{Roich-2021-TOG-PTI} in correcting distortion. 
However, we also find that in some instances, PTI~\cite{Roich-2021-TOG-PTI} performs better because the input faces in these cases have lower distortion levels, close to weak perspective projection. 
}

\begin{figure}
    \centering 
    \includegraphics[width=0.8\columnwidth]{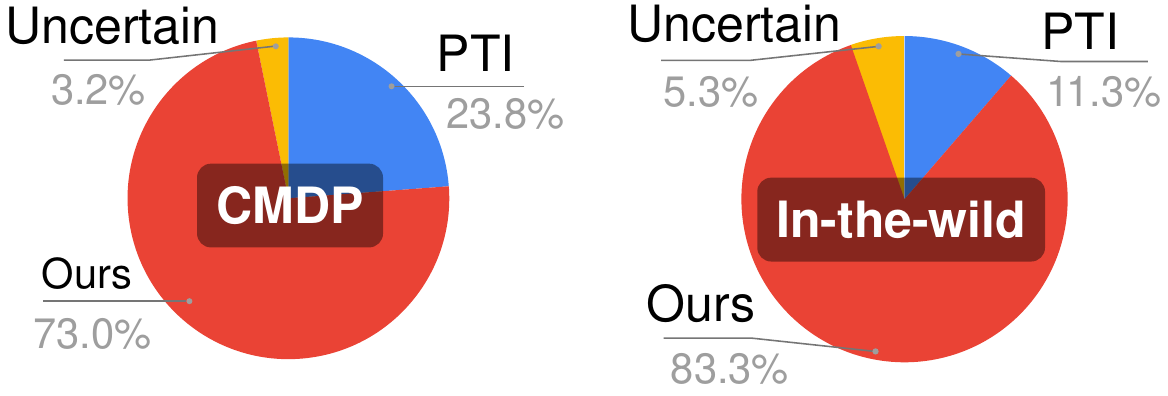}
    \caption{\revise{\tb{User study.} We conducted two user studies, one on the CMDP dataset~\cite{burgos2014distance} and another on our collected in-the-wild dataset. User prefer our results than PTI~\cite{Roich-2021-TOG-PTI}.}}
    \label{fig:user_study}
\end{figure}

\subsection{Bonus features}

Thanks to the generative ability of 3D GANs, our method enjoys additional advantages over warping-based methods in face completion and semantic editing.

\paragraph{Face completion}
\figref{part-face} demonstrates that our method can effectively correct the distortion in partially occluded faces. This capability is beneficial for seriously distorted faces near image boundaries, which cannot be handled by warping-based methods like \cite{fried2016perspective} due to the absence of face landmarks, or \cite{Zhao-2019-ICCV}, which cannot generate occluded regions.

\newlength{\qualfgw}
\setlength{\qualfgw}{0.242\linewidth}
 
\begin{figure}
    \centering
    \footnotesize
     \begin{tabular}
    {   
        @{\hspace{0mm}}c@{\hspace{0.5mm}} 
        @{\hspace{0mm}}c@{\hspace{0.5mm}}
        @{\hspace{0mm}}c@{\hspace{0.5mm}}
        @{\hspace{0mm}}c@{\hspace{0.5mm}}
    }
    \fbox{\includegraphics[width=\qualfgw]{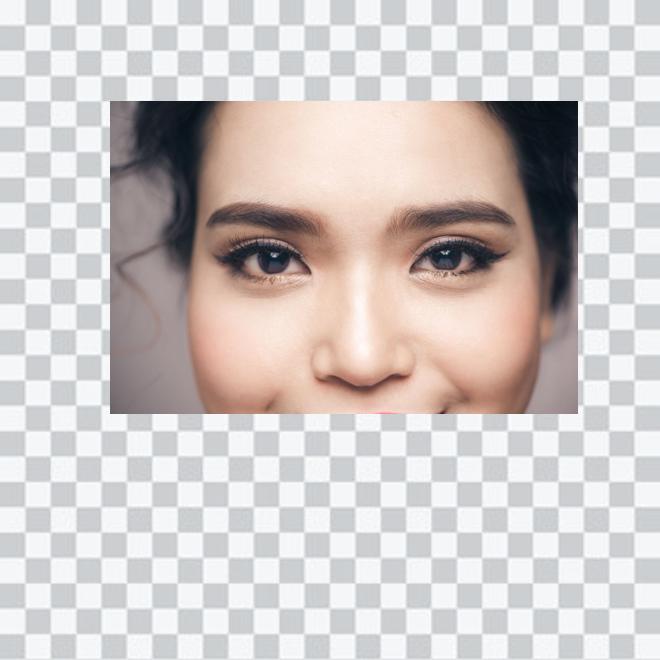}} &
    \fbox{\includegraphics[width=\qualfgw]{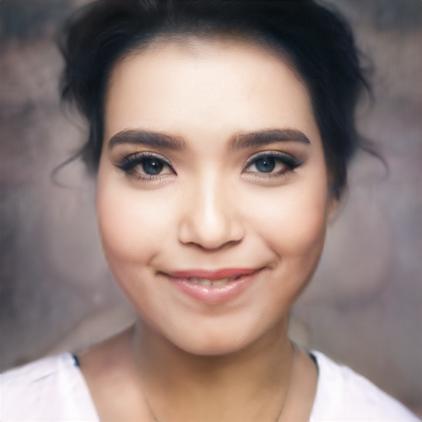}} &
    \fbox{\includegraphics[width=\qualfgw]{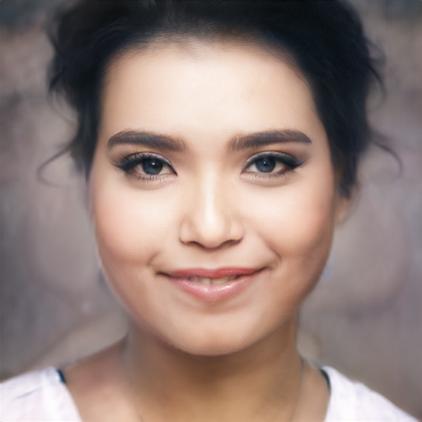}} &
    \fbox{\includegraphics[width=\qualfgw]{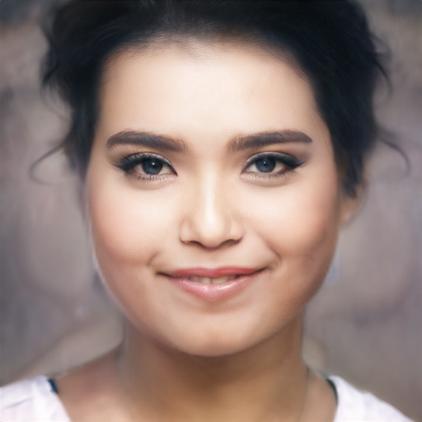}} \\
    \fbox{\includegraphics[width=\qualfgw]{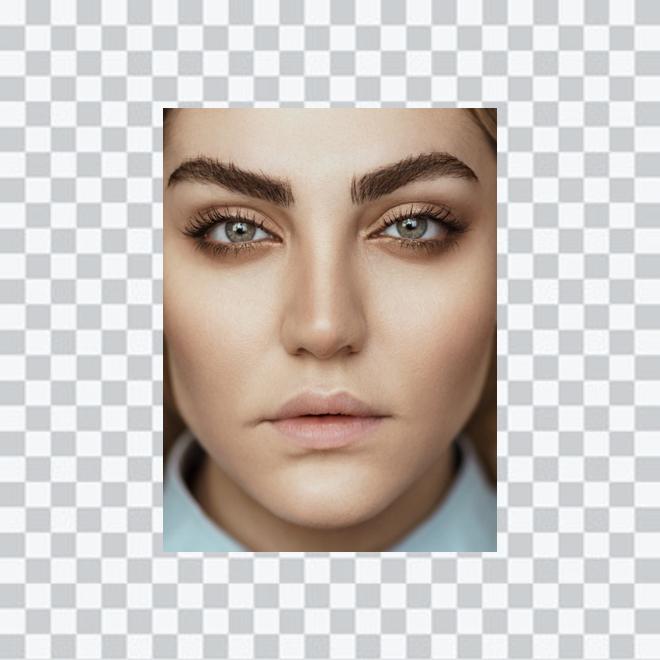}} &
    \fbox{\includegraphics[width=\qualfgw]{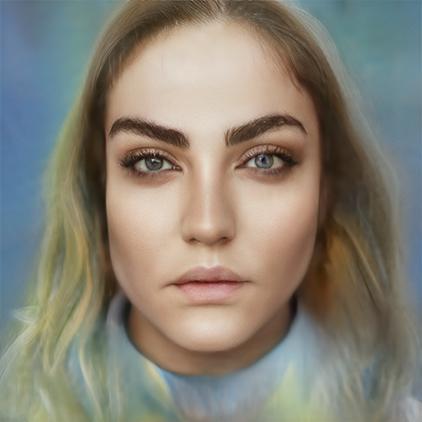}} &
    \fbox{\includegraphics[width=\qualfgw]{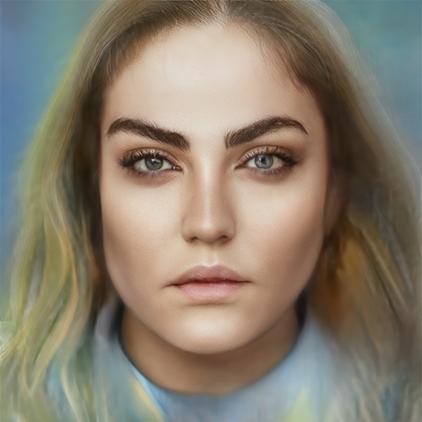}} &
    \fbox{\includegraphics[width=\qualfgw]{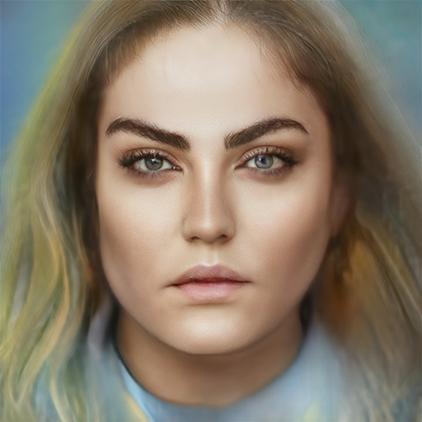}} \\
    Input &  Output $\times$1 & Output $\times$2  & Output $\times$4 \\
    \end{tabular}
    \vspace{-2mm}
    \caption{\textbf{Face completion.} Our method can apply directly to partially-occluded faces and does not expect a well-processed face. 
    }
    \label{fig:part-face}
\end{figure}

\paragraph{GAN editing}
\figref{edit} shows that our method improves the editing ability of 3D GAN on perspective-distorted input face images. 
Inverting the input distorted face with PTI~\cite{Roich-2021-TOG-PTI} can lead to an out-of-distribution facial latent code.
Editing these latent codes could generate unwanted artifacts.
Instead, our method inverts the image to an in-distribution face latent code that can be edited more accurately.

\begin{figure}
    \centering
    \fbox{\includegraphics[width=\columnwidth]{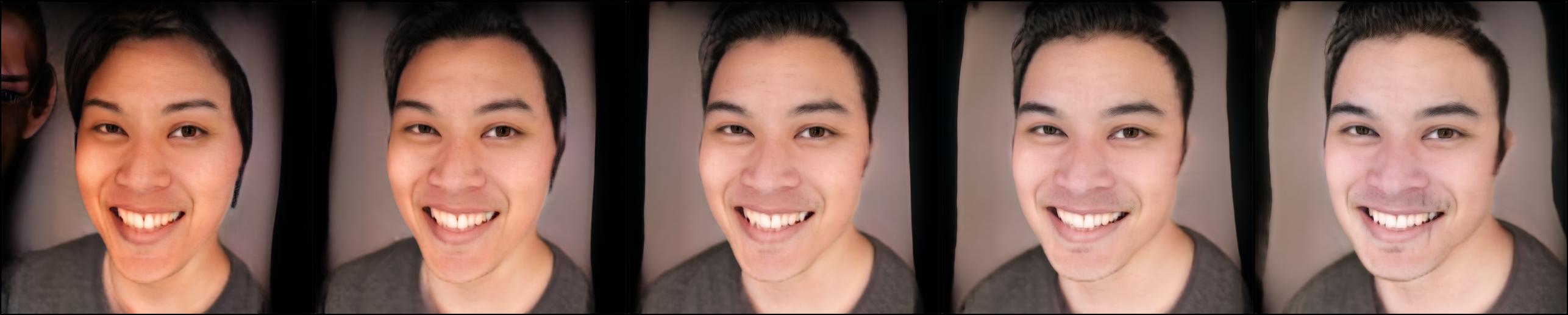}}\\
    \fbox{\includegraphics[width=\columnwidth]{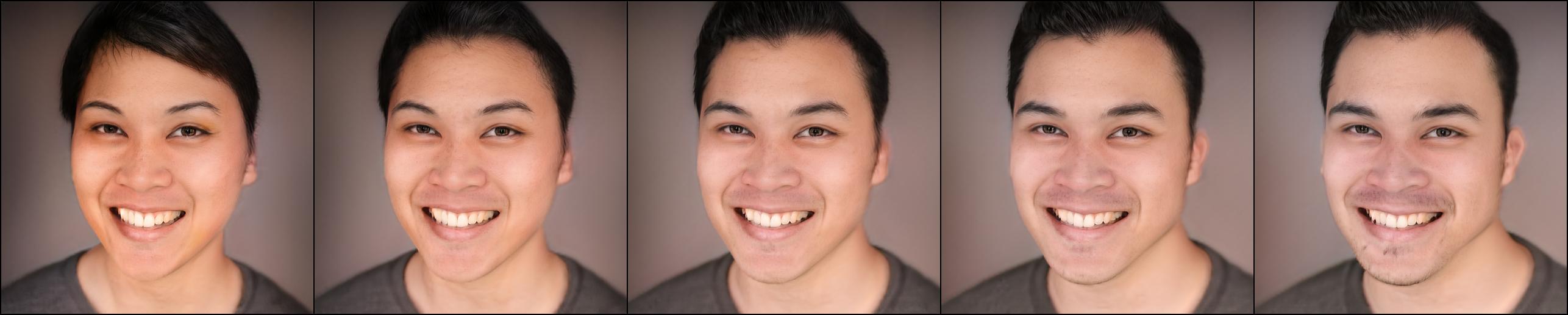}}\\
    \caption{\textbf{Editing ability.} Our method (\emph{bottom}) improves the editing ability of 3D GAN on perspective-distorted faces. Without our method (\emph{top}), inverting the input distorted face leads to an \emph{out-of-distribution} face latent code. Consequently, it leads to poor editing quality. 
    On the other hand, our method inverts an \emph{in-distribution} face latent code that enables us to edit. 
    It facilitates downstream applications. 
    }
    \label{fig:edit}
\end{figure}

\subsection{\revise{Limitations}}

\revise{While we advocate for our method, it has limitations, including its inability to handle out-of-distribution faces and its inability to process in real-time.}

\paragraph{Out-of-distribution faces}
As shown in \figref{failure},
our method fails for out-of-distribution faces, including extreme expressions and occluded faces (by hand or other objects).
In these cases, GAN inversion struggles to comprehend the face and may generate the face based on its own interpretation (e.g., the left example in \figref{failure} where the tongue is mistaken as part of the lip in the output). This can result in dreadful artifacts, as seen in the right example of \figref{failure}, where the hand looks distorted in the output.
\revise{A potential solution is first to mask these regions for GAN inversion. Then, transfer the textures to the manipulated face.}

\newlength{\limitw}
\setlength{\limitw}{0.213\columnwidth}
\newcommand{\limitrow}[1]{            
	\fbox{\includegraphics[height=\limitw]{limitation/#1_input.jpg}} &
	\fbox{\includegraphics[height=\limitw]{limitation/#1_ours.jpg}}
}
\begin{figure}
    \centering
    \footnotesize
     \begin{tabular}
    {   
        @{\hspace{0mm}}c@{\hspace{0.5mm}} 
        @{\hspace{0mm}}c@{\hspace{0.5mm}}
        @{\hspace{0mm}}c@{\hspace{0.5mm}}
        @{\hspace{0mm}}c@{\hspace{0.5mm}}
    }
    \limitrow{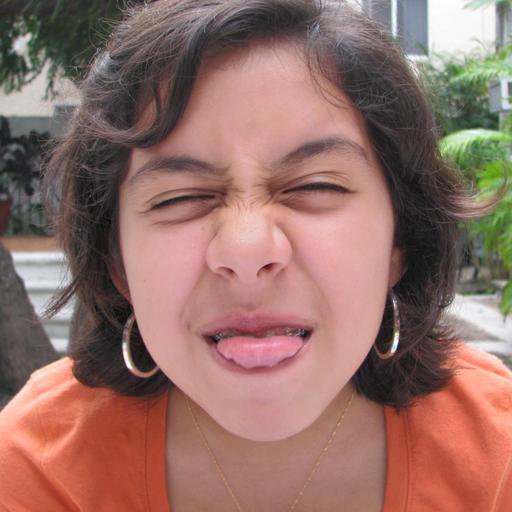} & \limitrow{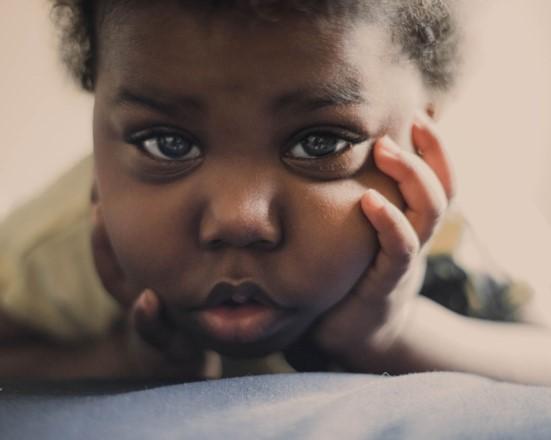}\\
    Input & Inversion & Input & Inversion \\
    \end{tabular}
    \vspace{-2mm}
    \caption{\textbf{Failure cases.} Limited by the training set of GAN, our method cannot handle out-of-distribution faces, \eg, tongue outside the mouth (\emph{left}), hand touch face (\emph{right}). A potential solution is first to mask these regions for GAN inversion. Then, transfer the textures to the manipulated face.
    }
    \label{fig:failure}
\end{figure}

\revise{
\paragraph{Inference speed}
We recognize that the current system 
does not operate in real time. 
Specifically, the GAN inversion process takes approximately 130 seconds to process a cropped face. 
This is because we implement our method based on the optimization-based inversion.
The time required for optimization is in line with PTI~\cite{Roich-2021-TOG-PTI}.
However, recent advancements~\cite{trevithick2023real,yuan2023make,bhattarai2024triplanenet} explored \emph{encoder-based} inversions for 3D GANs have successfully reduced inference times to less than 1 second.
These methods hold the potential to be seamlessly integrated into our perspective-aware 3D GAN inversion, significantly enhancing inference speed. Additionally, the encoder-based approach can overcome our current limitation of optimizing each individual photo.
Applying these encoder-based methods to our task would require training the encoder with paired perspective-distorted and ground-truth undistorted images. We leave the extension of speed improvement to future work. 
}

\section{Conclusions}
\label{sec:conc}
We present a method for portrait perspective distortion correction.
Our core idea is to leverage a 3D GAN inversion method to recover plausible facial geometry and reveal hidden facial parts such as ears.
We explore several design choices such as closeup camera-to-face distance initialization, optimization scheduling, focal length reparameterization, and landmark constraints.
Furthermore, we establish a protocol of quantitative evaluation for the portrait perspective distortion correction.
Quantitative and visual comparisons demonstrate the improved performance of our pipeline over existing methods.

{\small
\bibliographystyle{ieee_fullname}
\bibliography{egbib}
}
\newpage

\newpage
\section*{Appendix}
\appendix

\section{Discussions}

\subsection{Comparison with existing GAN inversion}

\paragraph{3D GAN inversion for far vs. close-up portraits (\figref{weak_perspective})}
Existing 3D GAN inversion methods \cite{sun2022ide,ko20233d,lin20223d} are designed for input face images captured at far distances, where the weak perspective model can be approximated, and inversion is easier due to the reflection of ground truth faces. Therefore, they may use inaccurate camera-to-subjective distances and focal lengths. However, our method targets perspective undistortion and is meant for close-up face images. In this scenario, the face latent code with different camera-to-subjective distances and focal lengths can generate faces with significant variations. As a result, estimating accurate camera-to-subjective distance and focal length becomes crucial for producing high-quality 3D face images.

\begin{figure}[bh]
    \centering
    \includegraphics[width=\columnwidth]{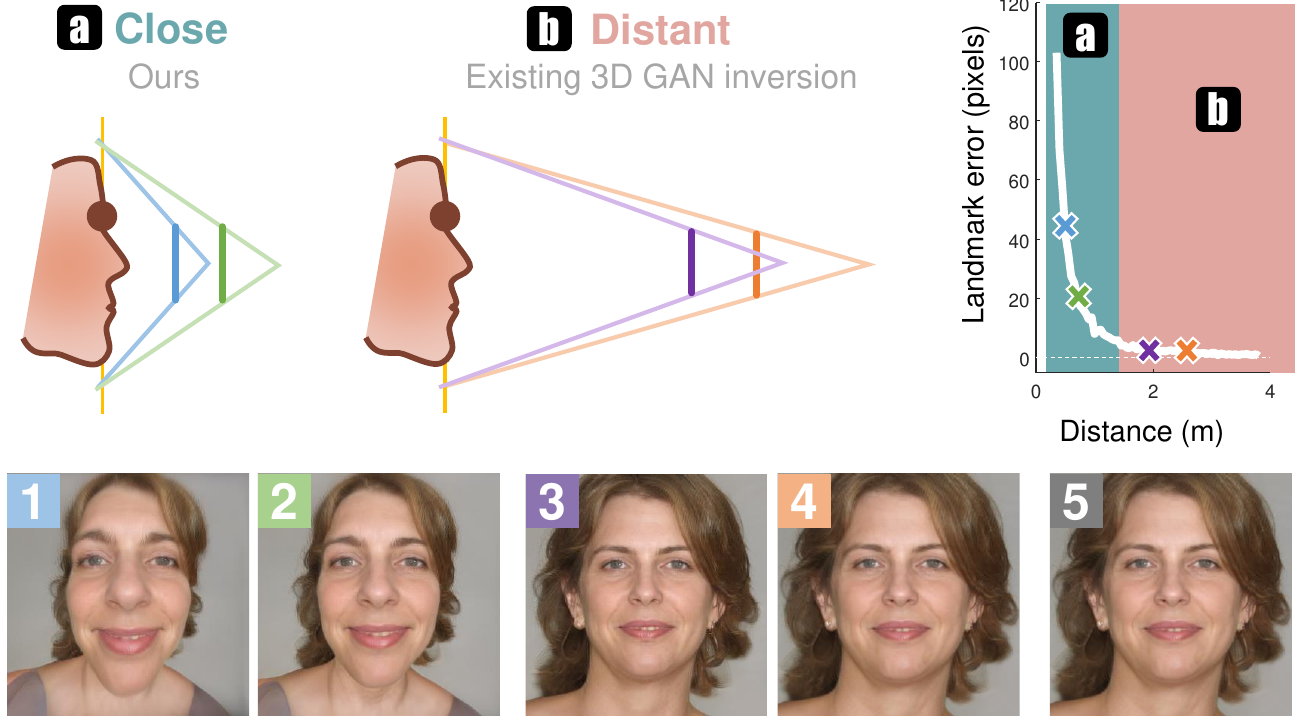}
    \caption{Our perspective-aware GAN inversion method differs from conventional GAN inversion approaches as it specifically focuses on close distances (\textbf{a}), whereas existing methods like \cite{sun2022ide,ko20233d,lin20223d} target far distances where a weak perspective model can be reasonably approximated (\textbf{b}). By comparing landmark errors between face images rendered with various camera parameters and the corresponding ground truth face, we observe that the error decreases exponentially as the imaging distance increases. We observe that the distance between images 1 and 2 is similar to that between 3 and 4. However, the faces in images 1 and 2 exhibit significant differences, while the faces in images 3 and 4 appear similar. Additionally, images 1 and 2 show distinct variations from the ground truth image 5, while images 3 and 4 share similarities with it.
    \vspace{-2mm}
    }
    \label{fig:weak_perspective}
\end{figure}

\paragraph{Comparison to PTI~\cite{Roich-2021-TOG-PTI}}

We find that in certain cases in \figref{user_study}, PTI performs better, especially when the input face has lower distortion levels, close to weak perspective projection. 
In \figref{user}, we utilize synthetic data to reveal that as the distortion level decreases, the performance difference between the two methods also diminishes.

\begin{figure}[!h]
    \centering 
    \vspace{-2mm}
    \includegraphics[width=\columnwidth]{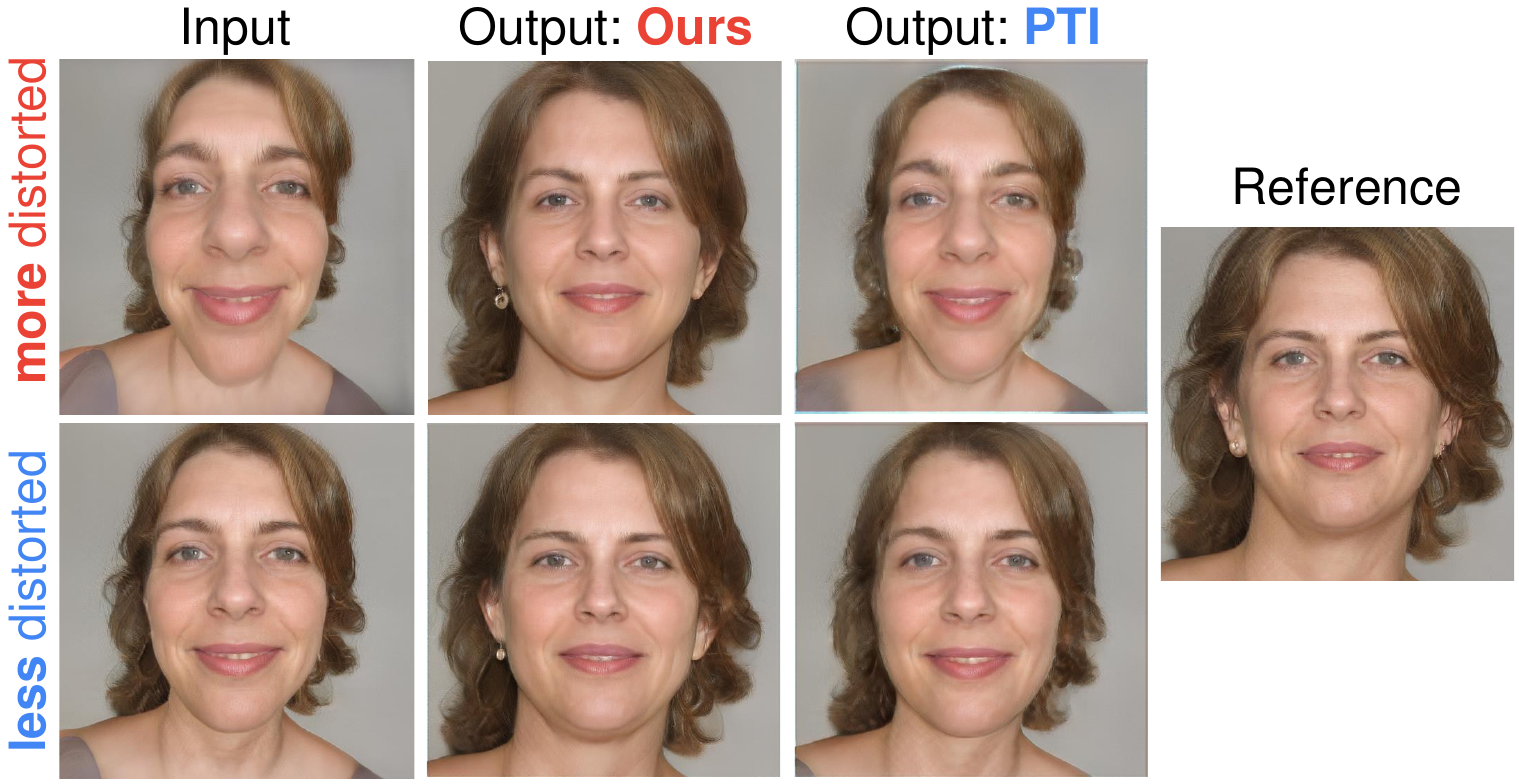}
    \caption{{Comparison with PTI on synthetic data.}}
    \label{fig:user}
\end{figure}

\paragraph{Visualization of inversion process}
In \figref{av}, we visualize the optimization process. We observe that without our perspective-aware designs, 3D GAN inversions often get trapped in local minima and fail to reconstruct the correct face geometry or correct the perspective distortion. Our proposed method overcomes these limitations and produces more accurate geometries and visually pleasing results.

\newlength{\qualcw}
\setlength{\qualcw}{0.12\textwidth}
\newcommand{\qualcrow}[5]{            
	\fbox{\includegraphics[width=\qualcw]{supp/2#1.jpg}} &
         \fbox{\includegraphics[width=\qualcw]{supp/5#1.jpg}} &
        \fbox{\includegraphics[width=\qualcw]{supp/10#1.jpg}} &
        \fbox{\includegraphics[width=\qualcw]{supp/12#1.jpg}} &
        \fbox{\includegraphics[width=\qualcw]{supp/18#1.jpg}} &
        \fbox{\includegraphics[width=\qualcw]{supp/551#1.jpg}} &
        \fbox{\includegraphics[width=\qualcw]{supp/tp0-MSBXWE_#1.jpg}} &
	{\includegraphics[trim={#2 #3 #4 #5}, clip,width=\qualcw]{supp/tp0-mesh-#1.jpg}}
}
\begin{figure*}[!h]
    \centering
    \footnotesize
     \begin{tabular}
    {   
        @{\hspace{0mm}}c@{\hspace{0.5mm}} 
        @{\hspace{0mm}}c@{\hspace{0.5mm}}
        @{\hspace{0mm}}c@{\hspace{0.5mm}}
        @{\hspace{0mm}}c@{\hspace{0.5mm}} 
        @{\hspace{0mm}}c@{\hspace{0.5mm}} 
        @{\hspace{0mm}}c@{\hspace{0.5mm}} 
        @{\hspace{0mm}}c@{\hspace{0.5mm}} 
        @{\hspace{0mm}}c@{\hspace{0.5mm}} 
    }
    \qualcrow{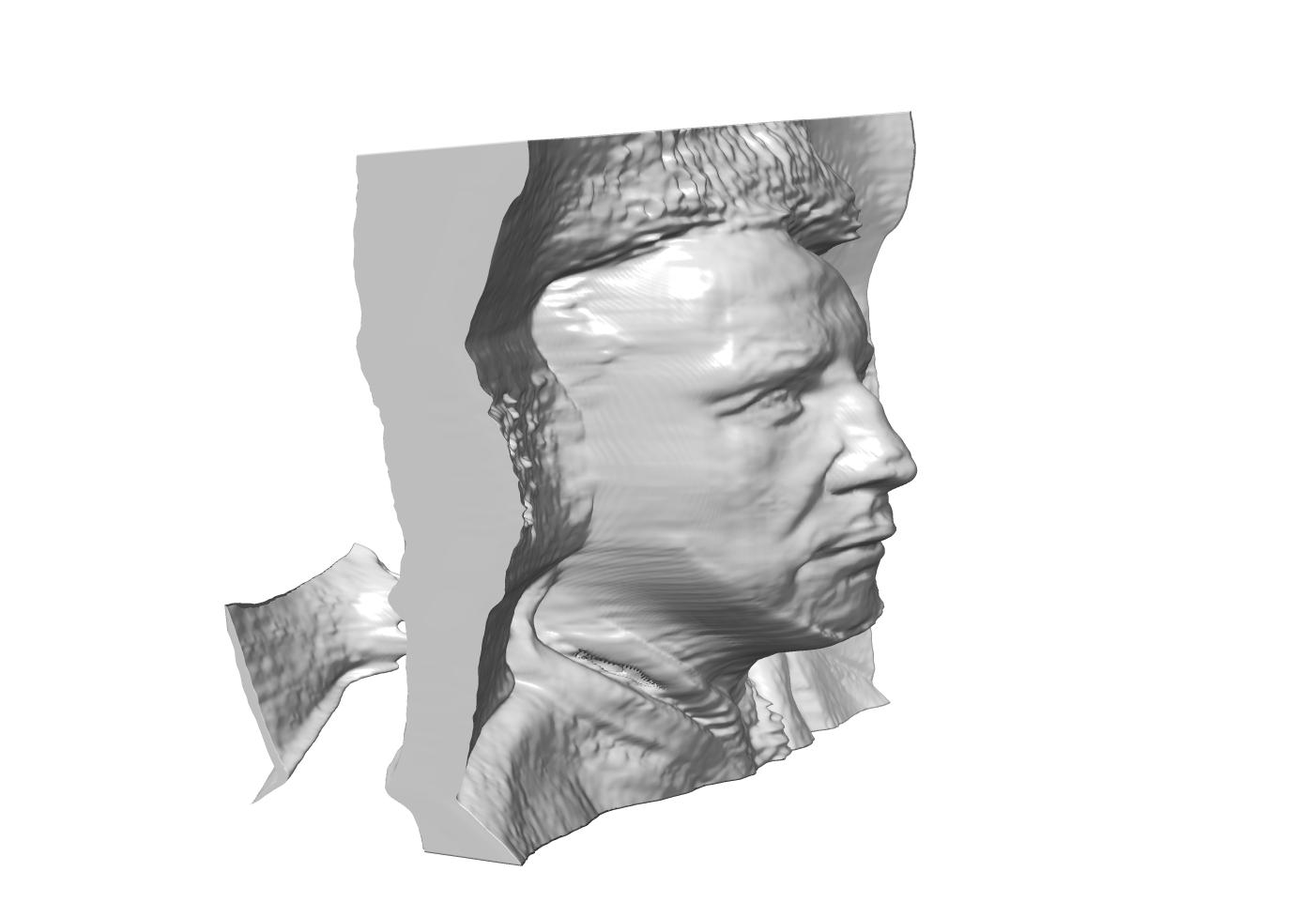}{5cm}{1cm}{7cm}{2.5cm}\\
    \qualcrow{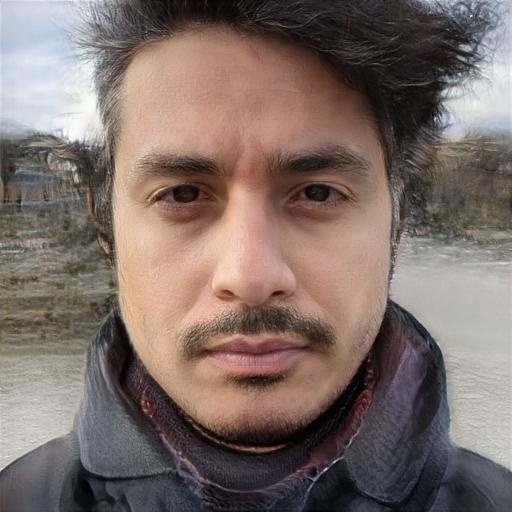}{5cm}{1cm}{7cm}{2.5cm}\\
    \qualcrow{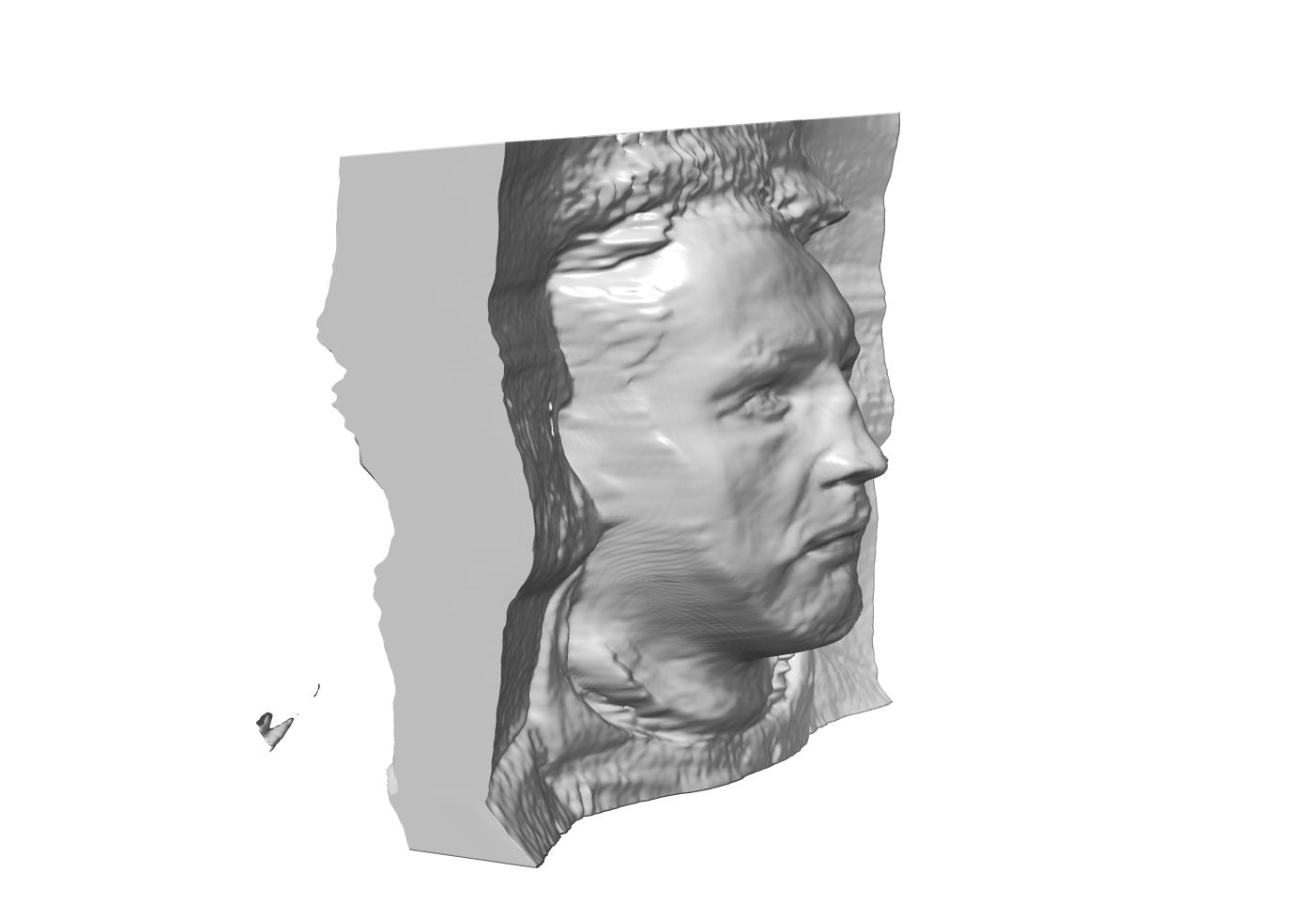}{5cm}{1cm}{7cm}{2.5cm}\\
    \setarial{Step: 20} & \setarial{Step: 100} & \setarial{Step: 200} &\setarial{Step: 240} & \setarial{Step: 360} & \setarial{Final} & \setarial{Reprojection} & \setarial{Mesh}\\\
    \end{tabular}
    \vspace{-2mm}
    \caption{Visualization of optimization. Our method (\emph{bottom}) first optimizes the camera-to-subject distance and then the face latent code. In contrast, PTI~\cite{Roich-2021-TOG-PTI} (\emph{top}) and Ko~\etal~\cite{ko20233d}  (\emph{middle}) optimize the face latent code while maintaining a fixed, incorrect camera-to-subject distance. This approach makes them susceptible to local minima, resulting in inaccurate shapes, such as those lacking ears.
    }
    \label{fig:av}
\end{figure*}

\subsection{Motivation for method design}

\paragraph{Alleviating ambiguity}
Reconstructing the correct face geometry from distorted images for perspective undistortion relies on accurately estimated camera parameters. To address this challenge, we propose a joint optimization approach that considers both face and camera parameters. However, the ambiguity in \figref{ambiguity} makes the task challenging. As shown in \figref{naiive}, adding na\"ive camera optimization with PTI does not yield satisfactory results. To overcome this, we design a perspective-aware inversion method that effectively alleviates ambiguity.
\begin{figure}[!h]
    \centering
    \includegraphics[width=\columnwidth]{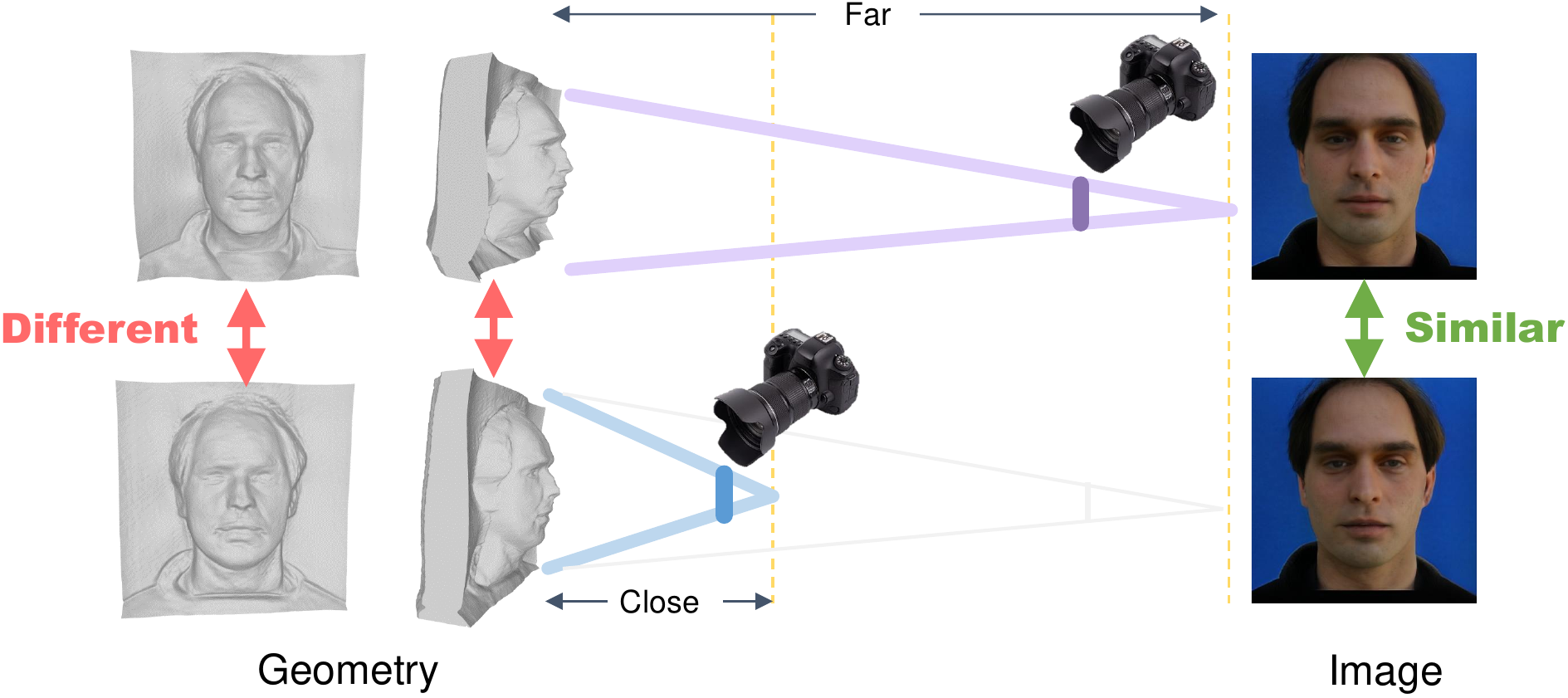}
    \caption{The ambiguity problem arises from the fact that multiple combinations of focal length, camera-to-subjective distance, and face shapes can result in similar faces. Consequently, if the camera parameters are estimated incorrectly, it can lead to incorrect face geometry for a given image.}
    \label{fig:ambiguity}
\end{figure}

\begin{figure}[!h]
    \centering
    \includegraphics[width=\columnwidth]{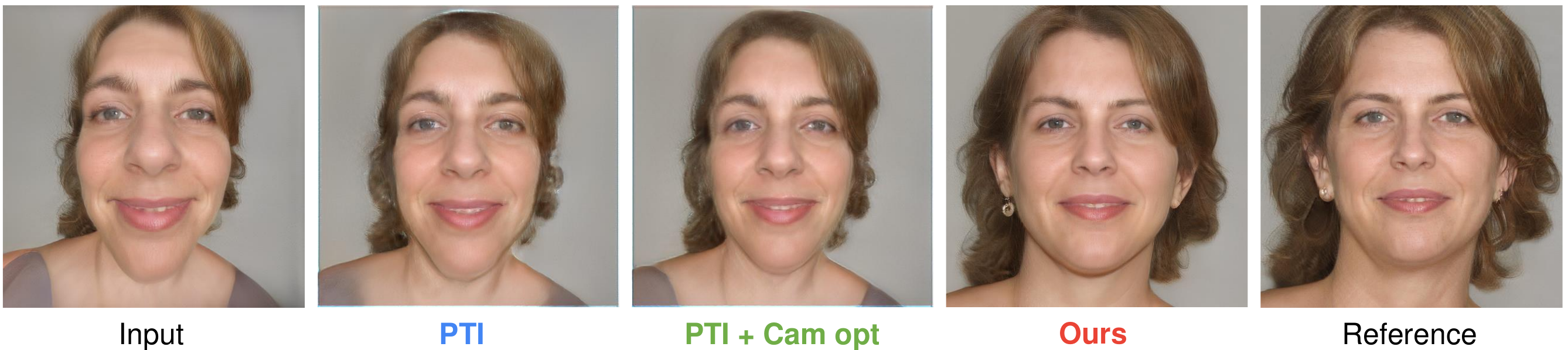}
    \caption{Na\"ive camera optimization with PTI does not provide significant improvement; in fact, its performance is similar to PTI alone.}
    \label{fig:naiive}
\end{figure}

\paragraph{Optimization scheduling}
When the camera parameters are incorrect, the face optimization process is more likely to fall into local minima, which in turn leads to the failure of camera parameter optimization. This interdependence between face and camera optimization makes the problem particularly challenging. Hence, we propose to optimize camera firstly.

\paragraph{Focal length reparameterization}

The reparameterization is motivated by two reasons. (1) During camera optimization, we observe that the focal length is more sensitive than the camera-to-subjective distance, making it difficult to optimize the latter. (2) Focal length and camera-to-subjective distance are related, and adjusting the focal length when changing the distance allows us to maintain the same FOV, reducing the degree of freedom in optimization.

\begin{figure*}[!h]
    \centering
    \includegraphics[width=0.9\textwidth]{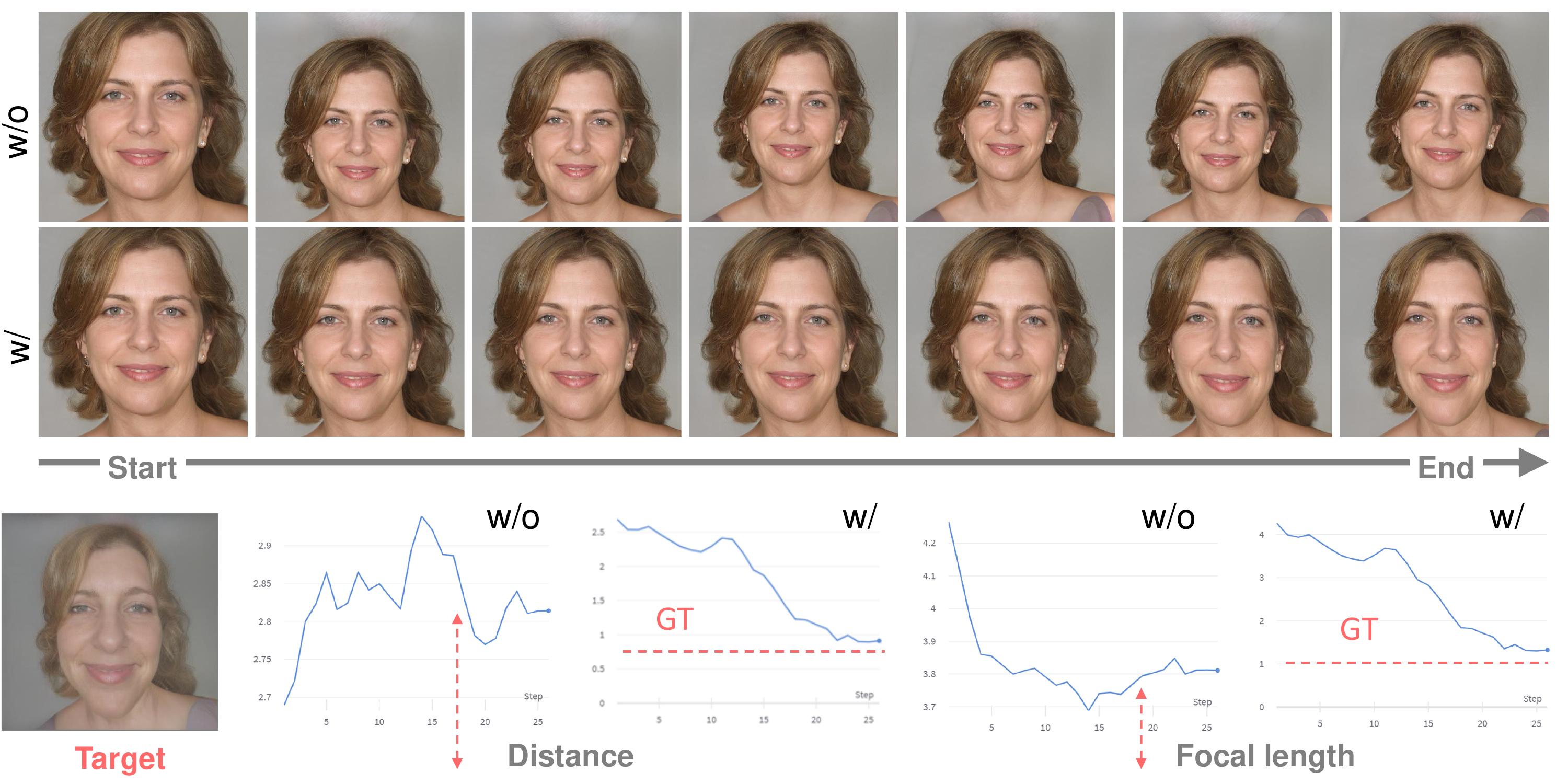}
    \caption{The optimization of camera-to-subjective distance can be challenging. To demonstrate this, we use a target image rendered by our 3D GAN and find the camera parameters using its ground truth face latent code. Without focal length reparameterization (w/o), adjusting the distance becomes difficult. However, with our focal length reparameterization (w/), optimizing the distance and approaching the ground truth (GT) distance becomes easier.}
    \label{fig:dist_opt}
\end{figure*}

\section{Method Details}
\label{sec:imp}

\subsection{Derivation of \eqnref{alpha}}

Let $\mathbf{p}_{c0} = \left( X_0,Y_0,Z_0 \right)^T \in \mathds{R}^{3}$ denotes the initial coordinate of one eye in the camera system. Its corresponding coordinate in the world system is given by
\begin{equation}
    \mathbf{p}_{w0} = \mathbf{R}_0^{-1}\left( \mathbf{p}_{c0}-t_0 \right)\,.
\end{equation}
Changing the camera to $\mathbf{R}, \mathbf{t}$ yields a new coordinate
\begin{equation}
\mathbf{p}_{c} = \mathbf{R}\mathbf{R}_0^{-1}\left( \mathbf{p}_{c0}-t_0 \right) +\mathbf{t}\,,
\end{equation}
where $\mathbf{p}_{c} = \left( X,Y,Z \right)^T \in \mathds{R}^{3}$, and $Z$ is equivalent the camera-to-subjective distance $d$.
We assume the rotation matrix changes slightly, \ie, $\mathbf{R}_0 \!\approx\! \mathbf{R}$. Hence, we have 
\begin{equation}\label{eq:approx2}
\mathbf{p}_{c} \approx \mathbf{p}_{c0} - t_0 + \mathbf{t}\,,
\end{equation}
We also assume $t_x$, $t_y$, $c_x$, and $c_y$ do not change.
To guarantee the eye position is fixed, we have the relationship $f/f_0 \!=\! d/d_0 \!=\! Z / Z_0 \!=\! \alpha\,$.
Substituting \eqnref{approx2} into the relationship, we obtain  the solution:
\begin{equation}\label{eq:alpha2}
    \alpha = (d_0 - (t_{z0} - t_z)) / d_0 \,.
\end{equation}

\subsection{Algorithm of perspective-aware 3D GAN inversion}

\begin{algorithm}
  \footnotesize
  \SetAlgoLined
  \KwIn{Pre-trained generator $G_\theta$.}
  \KwOut{Optimized camera parameter $\hat{\mathbf{c}}$, face latent code $\hat{\mathbf{w}}$, generator $G_\vartheta$, and updated parameters $d_0$, $f_0$ and $t_{z0}$.}
  \vspace{0.2cm}
  
  \textcolor{nice-green}{// Initialization} \\
  Get camera parameters $\mathbf{c}_0$ with focal length $f_0$ and $z$-axis translation $t_{z0}$.\\
  Get the face latent code $\mathbf{w}_0$.\\
  Get the camera-to-face distance $d_0$.\\
  Initialize $\mathbf{c} \leftarrow \mathbf{c}_0$, $\mathbf{w} \leftarrow \mathbf{w}_0$, $\delta t_z\leftarrow 1$, $ \gamma \leftarrow 1$.\\
  Get a close-up distance $\mathbf{t} \leftarrow \epsilon$.\\
  Get $\alpha$ according to Equation~(\textcolor{nice-red}{8}).\\
  Update $f \leftarrow \alpha f_0$.\\
  \vspace{0.2cm}
  
  \textcolor{nice-green}{// Optimize camera parameters}\\
  Fix face latent code $\mathbf{w}$, weights of $G_\theta$.\\
  \While{ \emph{iterations} $k < 300$}{
  Get the gradients $\nabla_\mathbf{t}$, $\nabla_\mathbf{R}$,  $\nabla_{\gamma}$.\\
  Optimize $\delta t_z \leftarrow \delta t_z + \lambda_\text{cam}\nabla_\mathbf{t}$.\\
  Optimize $t_z \leftarrow {t_{z0}}/{\sqrt{\delta t_z}}$.\\
  Get $\alpha$ according to Equation~(\textcolor{nice-red}{8}).\\
  Update $f \leftarrow \gamma \alpha f_0$.\\
  Optimize $\mathbf{p}\leftarrow \mathbf{p} + \lambda_\text{tiny}\!\times\!\lambda_\text{cam}\nabla_\mathbf{p}$,\, $\mathbf{p}\in \{\mathbf{R},t_x, t_y, \gamma \}$.
  }
  \vspace{0.2cm}
  
  \textcolor{nice-green}{// Optimize camera and face parameters}\\
  Fix weights of $G_\theta$.\\
  \While{ \emph{iterations} $k < 700$}{
  Get the gradients $\nabla_{\mathbf{t}}$, $\nabla_\mathbf{R}$, $\nabla_\mathbf{w}$,  $\nabla_\gamma$.\\
  Optimize $\delta t_z \leftarrow \delta t_z + \lambda_\text{cam}\nabla_\mathbf{t}$.\\
  Optimize $t_z \leftarrow {t_{z0}}/{\sqrt{\delta t_z}}$.\\
  Optimize $\mathbf{w} \leftarrow \mathbf{w} + \lambda_\text{face}\nabla_\mathbf{w}$.\\
  
  Get $\alpha$ according to Equation~(\textcolor{nice-red}{6}).\\
  Update $f \leftarrow \gamma \alpha f_0$.\\
  Optimize $\mathbf{p}\leftarrow \mathbf{p} + \lambda_\text{tiny}\!\times\!\lambda_\text{cam}\nabla_\mathbf{p}$,\, $\mathbf{p}\in \{\mathbf{R},t_x, t_y, \gamma \}$.
  }
  \vspace{0.2cm}
  
  \textcolor{nice-green}{// Pivotal tuning}\\
  Fix face latent code $\mathbf{w}$, camera parameters $\mathbf{c}$.\\
  \While{ \emph{not converge}}{
  Get the gradients $\nabla_\theta$.\\
  Optimize $G_\vartheta \leftarrow G_\theta + \lambda_\text{gan}\nabla_\mathbf{\theta}$.\\
  }
  Update  $\hat{\mathbf{c}} \leftarrow \mathbf{c}$, $\hat{\mathbf{w}} \leftarrow \mathbf{w}$\\
 Get $d$\\
  Update $d_0 \leftarrow d$, $f_0 \leftarrow f$, $t_{z0} \leftarrow t_z$
  \caption{Algorithm of perspective-aware 3D GAN inversion}
  \label{alg:init}
\end{algorithm}

\subsection{The proposed workflow}

\begin{itemize}
    \item \emph{3D GAN}: In our experiments, we employ the EG3D model~\cite{chan2022efficient} pre-trained on the FFHQ dataset~\cite{karras2019style}. Our method, however, is agnostic to the underlining 3D GAN models. For example, other 3D GANs such as IDE-3D~\cite{sun2022ide} could also be used. 
    \item \emph{Camera initialization}: We initialize the camera parameters by fitting a 3DMM~\cite{deng2019accurate}, consistent with the EG3D training process, ensuring the compatibility between the initialized camera parameters and EG3D. 
    \item \emph{Monocular depth estimation}: We incorporate the MiDaS approach~\cite{ranftl2020towards}.
    \item \emph{Reprojection}: we employ 3D Photo Inpainting~\cite{Shih3DP20} to reproject the background, including partial body and hair elements.
    \item \emph{Background inpainting}: As 3D Photo Inpainting~\cite{Shih3DP20} may not sufficiently reveal the hidden background and could result in undesirable gaps, we first use Stable Diffusion~\cite{rombach2021highresolution} or DALL$\cdot$E2 to inpaint the background when processing full-frame input images. 
    We then reproject the inpainted background and utilize it to replace the background in our rendered full-frame image. 
    For this task, we leverage MODNet~\cite{ke2022MODNet} to separate the person from the background.
\end{itemize}

\subsection{Parameters setting} 

\begin{itemize}
    \item We set learning rates: 
    \begin{itemize}
    \item $\lambda_\text{face}= 1\!\times\!10^{-2}$ 
    \item $\lambda_\text{face} = 5\!\times\!10^{-3}$ 
    \item  $\lambda_\text{gan} = 3\!\times\!10^{-4}$
    \item  $\lambda_\text{tiny} = 0.1$
    \end{itemize}
    \item We let the parameter $\epsilon$ equal 0.5
    \item We set the rendering parameters \texttt{ray\_start} and \texttt{ray\_end} to \texttt{auto} for close-up faces 
\end{itemize}

\section{Data Avability}
We evaluate our methods using three different datasets:
\begin{itemize}
    \item \textbf{Caltech Multi-Distance Portraits (CMDP) Dataset \cite{fried2016perspective}:}
This dataset is publicly available and has been referenced in our manuscript.

    \item \textbf{USC Perspective Portrait Database \cite{Zhao-2019-ICCV}:}
The USC perspective portrait database was collected by \cite{Zhao-2019-ICCV} from the internet.

    \item  \textbf{In-the-Wild Images:}
We collected in-the-wild testing images from the internet, such as \href{https://unsplash.com/license}{Unsplash}  and Adobe Stock, with \href{https://stock.adobe.com/license-terms}{a Standard license}. We will provide links for each image.
\end{itemize}

\end{document}